\documentclass[5p,times]{elsarticle}

\usepackage{hyperref}
\usepackage{subcaption}
\usepackage{graphicx}
\usepackage{xurl}
\usepackage{nomencl}
\makenomenclature

\journal{Applied Thermal Engineering}

\begin{document}

\begin{frontmatter}

% here are a list of the previous titles so we can refer back to them:
% Filling building data gaps: Multidimensional context autoencoder approach for building energy data imputation
% Multidimensional context autoencoder approach for building energy data imputation

% Tentative titles:
% 1. Employ image inpainting techniques to fill missing gaps in energy data
% 2. Imputation of missing building energy data by applying image inpainting techniques
% 3. Building energy data imputation at different missing rates using image inpainting techniques

\title{Filling time-series gaps using image techniques: Multidimensional context autoencoder approach for building energy data imputation}
\author[inst1]{Chun Fu}
\author[inst1]{Matias Quintana}
\author[inst2]{Zoltan Nagy}
\author[inst1]{Clayton Miller\corref{cor1}}

\ead{clayton@nus.edu.sg}
\cortext[cor1]{Corresponding author}

\affiliation[inst1]{organization={Building and Urban Data Science (BUDS) Lab, Department of the Built Environment, College of Design and Engineering, National University of Singapore (NUS)},
            country={Singapore}}
\affiliation[inst2]{organization={Intelligent Environments Lab (IEL), Department of Civil, Architectural and Environmental Engineering, Cockrell School of Engineering, The University of Texas at Austin},
            state={TX},
            country={USA}}

\begin{abstract}
Building energy prediction and management has become increasingly important in recent decades, driven by the growth of Internet of Things (IoT) devices and the availability of more energy data. 
However, energy data is often collected from multiple sources and can be incomplete or inconsistent, which can hinder accurate predictions and management of energy systems and limit the usefulness of the data for decision-making and research. 
To address this issue, past studies have focused on imputing missing gaps in energy data, including random and continuous gaps.
One of the main challenges in this area is the lack of validation on a benchmark dataset with various building and meter types, making it difficult to accurately evaluate the performance of different imputation methods.
Another challenge is the lack of application of state-of-the-art imputation methods for missing gaps in energy data.
Contemporary image-inpainting methods, such as Partial Convolution (PConv), have been widely used in the computer vision domain and have demonstrated their effectiveness in dealing with complex missing patterns. 
Given that energy data often exhibits regular daily or weekly patterns, such methods could be leveraged to exploit the regularity of the data to learn underlying patterns and generate more accurate predictions for missing values.
To study whether energy data imputation can benefit from the image-based deep learning method, this study compared PConv, Convolutional neural networks (CNNs), and weekly persistence method using one of the biggest publicly available whole building energy datasets, consisting of 1479 power meters worldwide, as the benchmark.
The results show that, compared to the CNN with the raw time series (1D-CNN) and the weekly persistence method, neural network models with reshaped energy data with two dimensions reduced the Mean Squared Error (MSE) by 10\% to 30\%. 
The advanced deep learning method, Partial convolution (PConv), has further reduced the MSE by 20-30\% than 2D-CNN and stands out among all models.
Based on these results, this study demonstrates the potential applicability of time-series imaging in imputing energy data. 
The proposed imputation model has also been tested on a benchmark dataset with a range of meter types and sources, demonstrating its generalizability to include additional accessible energy datasets. This offers a scalable and effective solution for filling in missing energy data in both academic and industrial contexts.

\end{abstract}


\end{frontmatter}

%%% Nomenclature
\nomenclature{ML}{Machine Learning}
\nomenclature{IoT}{Internet of Things}
\nomenclature{CNN}{Convolutional Neural Network}
\nomenclature{MSE}{Mean Squared Error}
\nomenclature{PConv}{Partial Convolution}
\nomenclature{RNN}{Recurrent Neural Network}
\nomenclature{GEPIII}{Great Energy Predictor III}
\nomenclature{BDG}{Building Data Genome}
\nomenclature{GAN}{Generative Adversarial Network}
\nomenclature{FCNN}{Fully-Connected Neural Network}
\nomenclature{FDD}{Fault Detection and Diagnosis}

\printnomenclature

\newpage
\section{Introduction}
Machine learning (ML) has recently demonstrated great \newline promise for energy forecasting by providing accurate and reliable predictions of future energy demand and supply. As detailed by Kazmi et al. \cite{kazmi2023ten}, the field is evolving, driven by increased computing power, advancements in data handling, and innovative algorithmic methods, which together have significant potential to transform energy demand forecasting at both building and urban scales. 
The ML algorithms have shown remarkable potential in predicting energy consumption in buildings \cite{somu2021deep,pham2020predicting,olu2022building}, forecasting renewable energy output \cite{dolara2017weather,gensler2016deep,tian2022developing}, and estimating grid electricity demand \cite{hafeez2020novel, avalos2020comparative, hafeez2020electric}. 

By outperforming traditional statistical methods, ML models have demonstrated significant improvements in terms of accuracy and reliability \cite{paterakis2017deep, touzani2018gradient}. 
Moreover, ML also has been applied in building operations and energy management, such as cooling load forecasting-based optimization for chiller control \cite{wang2019cooling} and ML-based methods for optimally managing energy sources with power grids \cite{shams2021artificial}.
However, with these advancements brought by ML, practical applications often face challenges due to instability in data quality of building energy consumption, which can result in problems with forecasting performance.
Among the issues with data quality, missing data is a primary one that hinders the efficacy of ML models. 
These missing data can originate from a spectrum of causes, ranging from terminal equipment and operator error to malfunctions of sensors \cite{gao2015missing, sun2013robust}. 
The missing data can have cascading consequences, negatively impacting stability, performance, and failure prevention \cite{stones2001power}.
In certain commercial buildings, it can result in energy waste ranging from 15 and 30 percent \cite{khan2013fault, choi2021autoencoder}. 
Given the criticality of the issue, our study seeks to utilize state-of-the-art deep learning methods to bridge the missing data gaps, thereby contributing a new perspective to the building energy data imputation field.

\subsection{Data mining and modeling techniques for addressing data quality issues}
The presence of abnormal data in energy forecasting is a significant issue, as emphasized in the ASHRAE Great Energy Predictor III (GEPIII) competition on Kaggle \cite{miller2020ashrae}. Additionally, the abnormal energy-consumption behavior may also limit the predictability of ML on energy forecasting \cite{miller2022limitations}. To address these issues, several initiatives, such as the Large-scale Energy Anomaly Detection (LEAD) competition, provides a benchmark dataset for energy anomaly detection in commercial buildings \cite{gulati2022lead1}, and a tree-based ensemble model was proposed for scalable anomaly detection \cite{fu2022trimming}. 
Notable advancements in research include the development of a model for multivariate time series anomaly detection in energy consumption based on a graph attention mechanism \cite{zhang2022multivariate} and an enhanced fault detection method for building energy systems in the presence of missing data using expectation-maximization and Bayesian network \cite{wang2021fault}. 
Concurrently, unsupervised techniques have also been extensively utilized to discern patterns in energy usage through clustering and to identify anomalies in daily profiles of energy data \cite{park2019apples, park2020good, miller2015automated}.

The applications of ML in the building life cycle are primarily focused on the operation and maintenance phase \cite{ds_energy2021}.
In particular, specific use cases need more data in quantity and diversity and thus opted for data-driven methods to tackle this.
As for data quantity, generative models, specifically Generative Adversarial Networks (GANs), have been used for scenarios without sufficient historical data, requiring synthetic data.
These GAN-based models were first proposed as a way to learn the distribution of the input data and generate synthetic data that resembles the input data \cite{Goodfellow2014}.
In the built environment, these models have been used in power demand prediction \cite{yeEnergyBuildingsEvaluating2022}, building load profile \cite{Wang2020c}, fault detection and diagnosis (FDD) \cite{Yan2020}, and indoor thermal comfort \cite{comfortGAN} to generate synthetic complementary data for model training.
Another approach is to augment energy time-series data using masking noise injection for more efficient imputation of missing values \cite{liguori2023augmenting}. Several data augmentation methods that manipulate time series in terms of magnitude and temporal order have also been discussed \cite{iwana2021empirical, demir2021data}. These papers emphasize the importance of data augmentation in time series analysis and energy modeling, especially in the absence of sufficient data.

Other approaches circumvent the lack of available and diverse data by incorporating external data streams to aid in the data mining task.
Fu et al. \cite{fu2022using} used Google searches, in the form of Google trends, as additional input data for energy consumption forecasting.
Moreover, Sun et al. \cite{sun2022b} used building facade images to complement energy consumption and building metadata for energy efficiency rating prediction.
These models showcase the incorporation of external data streams of different sensing modalities and their potential aid.
However, while these approaches increase the performance of the downstream task, they do not address the data availability issues. 
Incorporating new data streams may potentially introduce additional availability issues.

\subsection{Imputation models in energy and building fields}
To enhance data availability, a plethora of methods have been proposed for data imputation, ranging from traditional statistical methods, regression models, and deep learning.
The types of missing data can be categorized as short-term missing (or random missing) and long-term missing (or continuous missing).
Short-term missing (or random missing) refers to missing data points that cover a relatively small amount of time (e.g., hourly or daily missing), while long-term missing (or continuous missing) refers to data gaps that cover a longer period of time (e.g., weekly or monthly missing in time series). 
Traditional data reconstruction methods can be traced back to linear or polynomial regression methods, which have proven effective and efficient in data imputation, especially for short-term missing data \cite{chong2016imputation, wang2021towards}. 
Some other supervised-learning regression methods based on lag features were often used to impute missing energy data gaps.
For example, linear regression, weighted K-nearest neighbors (kNN), support vector machines (SVM), and mean imputation was implemented for random missing values varying from 5\% to 20\% missing rate, demonstrating superior performance than zero or mean imputation \cite{chong2016imputation}. 
Another comparative study by Wang et al. stated that both statistical methods and machine learning regression models exhibit good performance for imputing short-term missing gaps within a one-day period in time series data\cite{wang2021towards}. 
However, these methods have demonstrated limited performance in imputing long-term missing data, which is commonly encountered in real-world scenarios. 
This limitation can be attributed to the reliance of statistical and regression models on nearby values, which are unavailable for long-term missing data.

To address the long-term missing data mentioned above, neural networks have been utilized for data imputation due to their advantage of learning context in unstructured datasets. 
For example, a method using an improved Convolutional Neural Network (CNN) that considers the correlation of power data from dimensions of time and space was proposed for filling missing data \cite{wu2021improved}.
Recurrent Neural Network (RNN) and Bi-directional Long Short Term Memory model (BI-LSTM) were proposed to impute energy data missing due to the temporal aspects of energy data \cite{lucbert2022time, ma2020bi}.
More recently, autoencoder has received growing attention among various frameworks of neural networks because of its self-reconstruction and feature extraction capabilities. 
For example, a study applied an autoencoder to reconstruct missing gaps in indoor environmental quality data, showcasing improved performance compared to classic polynomial interpolations \cite{liguori2021indoor}.
In another study, a 2D Convolutional Neural Network (2D-CNN) autoencoder was employed to address missing energy data considering the weekly periodicity of the data. The energy data was reshaped into a two-dimensional format, and the proposed autoencoder effectively imputed both random and continuous missing data \cite{wang2020intelligent}. 
The studies above demonstrate how autoencoders can be applied to data imputation in the built environment for both random and continuous missing. 
However, most of the studies have been conducted in the context of a limited number of buildings or a single power grid, leaving their applicability to a broader set of scenarios (e.g., unseen buildings or different sites) untested. 
Moreover, these neural networks belong to earlier or outdated frameworks, and some modern deep learning techniques that could yield even more accurate predictions have yet to be thoroughly investigated for data imputation in this field. 
In light of these observations, the development of modern deep learning frameworks considering generalizability in application holds great potential for advancing data imputation methods in the built environment.

\subsection{Image-based data imputation}
Missing data is a ubiquitous problem in various fields, such as audio signals repair \cite{esquef2006efficient, marafioti2019context}, and image inpainting in the computer vision field \cite{pathak2016context, yu2018generative}. 
Modern frameworks developed within these fields to address missing data issues hold the potential to revolutionize the way missing data is handled in the field of building energy. 
One such framework is partial convolutions, which was proven effective in image inpainting tasks involving regular (e.g., circles and rectangles) and irregular holes in images \cite{liu2018image}. 
Another promising framework is the diffusion model, which employs iterative diffusion processes to fill in missing data while preserving the smoothness and continuity of the original signal \cite{lugmayr2022repaint}.
However, these modern image-based techniques have yet to be widely applied to building energy data.  
There are a few reasons for this. 
The first reason is that energy data has traditionally been treated and imputed as time series. 
Consequently, past research has primarily relied on statistical methods for time series or RNNs to handle missing gaps. 
This means that the structured context and two-dimensional representation of energy data that inpainting methods can leverage have yet to be fully utilized.
Another factor may be the lack of sufficient energy data for training.
Past studies often utilized a limited number of meters or grids for developing and evaluating the imputation model.
In contrast, fields like image or audio processing typically have access to larger datasets, allowing for more extensive training and evaluation to verify the generalizability and effectiveness of the proposed methods. 
For instance, benchmark image datasets often consist of thousands to millions of image samples \cite{jam2021comprehensive}, which far exceed the scale of typical energy datasets that may have, at most, thousands of energy meters.
In response to these challenges, this research aims to transform conventional one-dimensional time-series representation of meter data into two-dimensional images (i.e., time of week by week). We propose to implement an advanced image inpainting framework to impute missing gaps with assistance from the structured context inherent in energy data.
In addition, this research will harness the largest publicly available building energy dataset, encompassing more than one thousand power meters, for comprehensive model training and validation.

\subsection{Research objectives and novelty}

\begin{figure*}[!htb]
\begin{center}
\includegraphics[width=0.6\textwidth, trim= 0cm 0cm 0cm 0cm,clip]{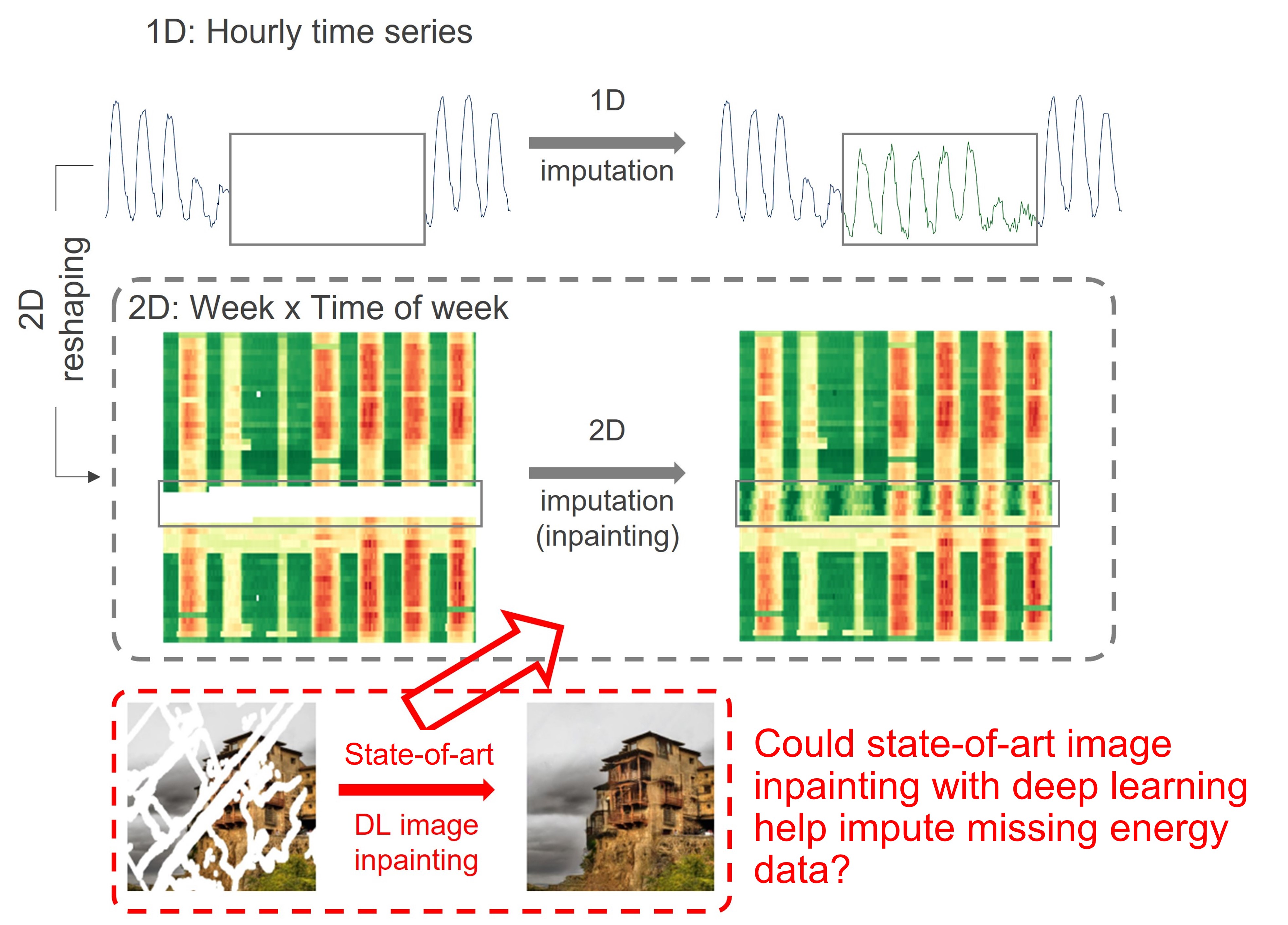}
\caption{The one-dimensional time series of energy data can be reshaped into a two-dimensional heatmap image for testing different imputation methods. Comparisons are made between the one- and two-dimensional imputation, including baseline model, 1D- and 2D-CNN, and advanced deep learning techniques used for image inpainting.}
\label{fig:overview}
\end{center}
\end{figure*}

To fill the aforementioned gaps, this study aims to investigate how the context of energy data affects data imputation performance. 
It achieves this through the use of a global dataset with diverse meters while setting different rates of both short-term and long-term missing data.
Leveraging deep learning techniques from the field of computer vision, this research reshapes time-series energy data to harness the cyclical nature of such data and applies state-of-the-art image-inpainting models for energy data imputation (as illustrated in Figure \ref{fig:overview}).
The research objectives of this study are to:
\begin{enumerate}
    \item Present a novel viewpoint in imputing energy data by converting time series into images and investigating image-based algorithms.
    \item Explore the use of modern deep learning frameworks, such as partial convolutions (PConv) and 2D-CNN, for imputing missing data in energy time series.
    \item Evaluate these frameworks' performance in imputing both random and continuous missing data with varying missing rates and test their generalizability to different sites.
    \item Benchmark the performance of different image-based frameworks (i.e., 2D-CNN and PConv) against baselines (i.e., 1D-CNN and weekly persistence method).
\end{enumerate}

\newpage
%Significance
This study's novelty lies in applying image-inpainting models, which have been widely used in the field of computer vision, to the imputation of energy data. 
This presents a new and innovative approach to the challenge of missing data in energy time series and offers the potential to further improve the accuracy of data imputation through the contextual learning of the underlying energy data structure. Consequently, it could enhance the data quality used for energy-related downstream tasks such as energy forecasting and building energy modeling. 
Furthermore, this study verifies the models' generalizability across various buildings and meter types and their effectiveness in addressing long-term missing data. 
By demonstrating the applicability and effectiveness of these models in the field of building energy, this study opens up new avenues for future research at the intersection of image-based techniques in the built environment.

\section{Methodology}
\label{sec:methods}
\begin{figure}[!htb]
\begin{center}
\includegraphics[width=0.45\textwidth, trim= 0cm 0cm 0cm 0cm,clip]{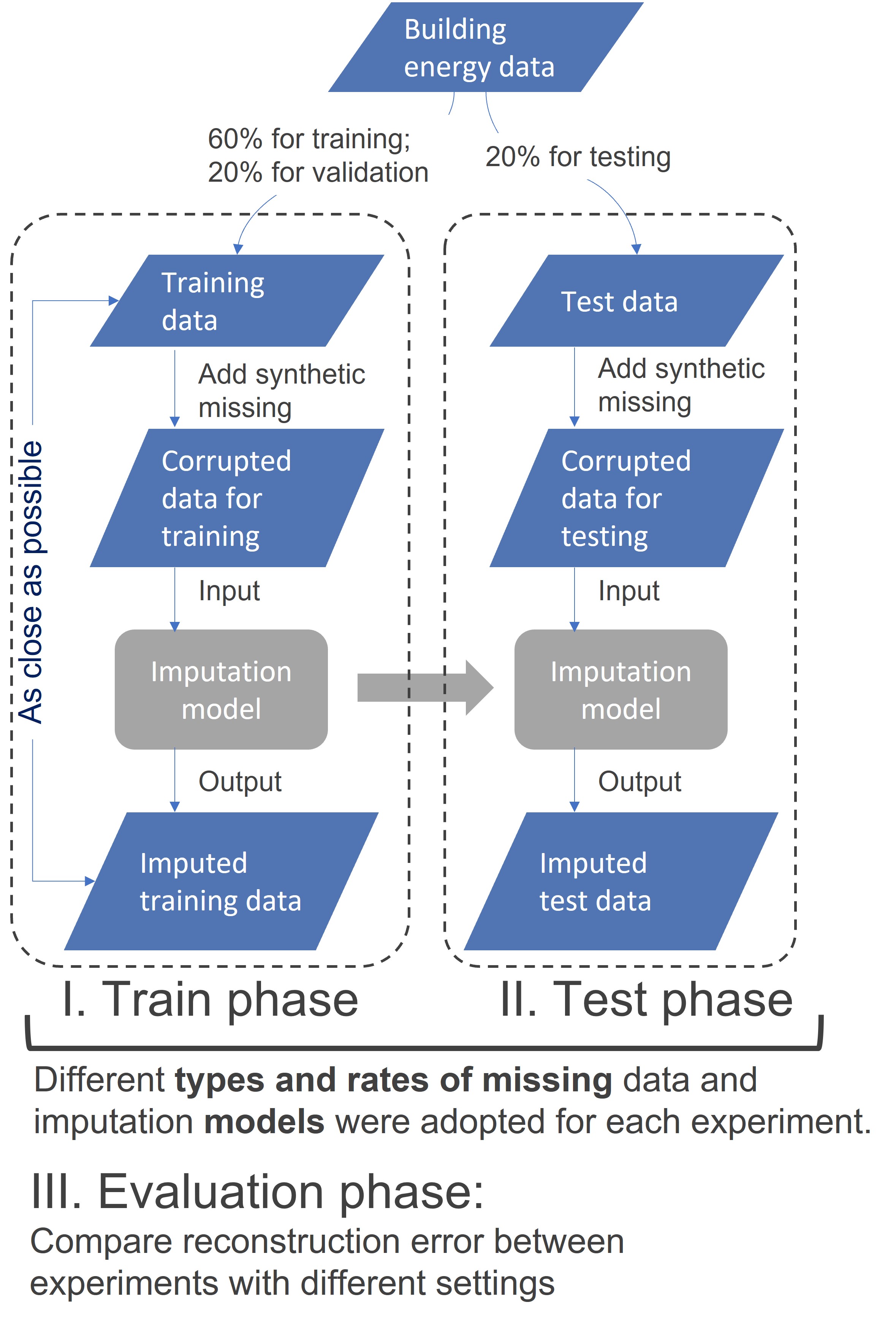}
\caption{Overview of the proposed framework and different phases in the working pipeline.}
\label{fig:workflow}
\end{center}
\end{figure}

The proposed methodology for this study consists of three phases, as illustrated in Figure \ref{fig:workflow}. 
In the first phase, training models for imputation will be developed using energy data with synthetic missing values as input and the corresponding raw data as output. 
The objective is to train a model that can accurately impute the synthetic missing values and produce results that closely resemble the raw data. 
In the second phase, the performance of these trained models will be evaluated and quantified on a test dataset with varying settings of missing and various meter types. 
Finally, in the third phase, a comprehensive comparison and discussion of the results obtained from all models, including the baseline method, will be conducted to draw conclusions.

\newpage
\subsection{Dataset: Building Data Genome 2.0 (BDG2)}
This study used the time-series hourly data from power meters in the Building Data Genome 2 (BDG2) project dataset for modeling. 
These data were also employed in the Great Energy Predictor III competition hosted on the Kaggle platform \cite{Miller2020-fo}. 
The Building Data Genome 2.0 (BDG2) is an open dataset containing hourly meter readings and metadata of 3,053 power meters over two years. 
Each building within the dataset is associated with metadata such as floor area, weather, and primary use type. 
Due to the variability of meter characteristics observed across thousands of meters worldwide, this dataset is an ideal benchmark dataset for comparing different machine learning algorithms while testing generalizability. 
Table \ref{tab:metadata} outlines the metadata variables available within the dataset.
It's worth noting that only the historical energy consumption data was used, and other potential data sources, such as weather and calendar data, were not incorporated in the imputation process. 
That's because the focus of this research was to impute missing data in energy time series by leveraging nearby values of the data.

\begin{table}[!htb]
\centering
\resizebox{0.47\textwidth}{!}{%
\begin{tabular}{llll}
\hline
\textbf{Category} & \textbf{Variable} & \textbf{Unit} & \textbf{Content} \\ \hline
\begin{tabular}[c]{@{}l@{}}Power\\ meter\end{tabular} & Meter type & - & \begin{tabular}[c]{@{}l@{}}Type of meter: electricity,\\ chilled   water, steam, or \\ hotwater\end{tabular} \\
 & Meter readings & kWh & Energy consumption \\
 &  &  &  \\
Building & Primary use & - & Primary category of activities \\
metadata & Year built & - & Year building was opened \\
 & Floors & - & Number of floors \\
 & Floor area & Sq foot & Gross floor area \\
 &  &  &  \\
Weather & Temperature & \textdegree C & Outdoor temperature \\
data & Cloud cover & Oktas & Portion of the sky covered \\
 & Dew point & \textdegree C & Outdoor dew temperature \\
 & Precipitation & Millimeter & Precipitation depth \\
 & Pressure & Millibar & Sea level pressure \\
 & Wind speed & m/s & Wind speed \\
 & Wind bearing & Degree & Wind direction \\ \hline
\end{tabular}
}
\caption{Building metadata available in the Building Data Genome 2 (BDG2) project.}
\label{tab:metadata}
\end{table}

\subsection{Data preprocessing}
The energy dataset will be subjected to several preprocessing steps to prepare it for model training. 
These steps encompass data cleaning to remove inconsistencies or errors, data normalization to standardize the data across a common range, data splitting into training, validation, and test sets, and data augmentation, such as flipping and shifting the time series to expand the dataset. 
By following these commonly used preprocessing steps in the domain of building energy data, the resulting models are equipped to deliver accurate and reliable predictions.

\subsubsection{Data cleaning}
Since this dataset is sourced from power meters installed in real-world buildings, it is expected to contain missing data resulting from system errors or equipment failures.
Cleaning the dataset by removing anomalies or errors is necessary to ensure the quality of the data used for training the imputation model. 
To accomplish this, we utilized the data cleaning results provided by the winning team in the GEPIII Kaggle competition, which has also been used as a benchmark for detecting anomalies in prior research \cite{quintana2022aldi++}. 
This process involves identifying and removing long streaks of constant values, large positive/negative spikes, and other anomalies determined through visual inspection.
After cleaning the dataset, we chose the power meters with low missing rates (less than 5\%) for our study, resulting in a total of 1479 meters (approximately 50\% of the entire BDG2 dataset).

\subsubsection{Data normalization}
To enhance the training process, the numerical values of the power meters were normalized to ensure consistent data distributions. 
Normalization is particularly essential when preparing the data for neural networks, as these models rely on the gradient descent algorithm to optimize their parameters. 
This algorithm can be affected by the scale of the data; thus, normalization helps achieve convergence. 
In this study, we applied min-max normalization, which scales the data to a range of $[0, 1]$. 
This is achieved by subtracting the minimum value in each series from the data and dividing by the difference between the maximum and minimum values (as shown in Equation \ref{eqn:normalization}).

\begin{equation}
\label{eqn:normalization}
X_{norm} = \frac{X-X_{min}}{X_{max}-X_{min}}
\end{equation}

\subsubsection{Data splitting}\label{method:data-split}
This study employed a 5-fold cross-validation technique to divide the 1479 power meter data into training, validation, and testing sets. 
Each round of modeling utilized 60\% of the data for training, 20\% for validation to improve training efficiency, and 20\% for testing to evaluate model performance. 
To ensure the model's generalizability, the data was divided by site ID to prevent the model from imputing missing data based on similar meters within the same site. 
This means the test dataset for evaluating the model's performance was from unseen sites for the trained model.
This allows us to check if the model is able to accurately impute data for meters from new sources, which is crucial in real-world scenarios.

\subsubsection{Data augmentation}
Data augmentation was employed in this study due to the small dataset size (1479 time series or images) available for developing machine learning models for meter-wise imputation, particularly for image-based Convolutional Neural Networks (CNNs) and modern image inpainting methods. 
To address this issue, the study utilized two specific data augmentation techniques for time-series data: shifting and flipping, as illustrated in Figure \ref{fig:missing_type}. Shifting moves the time series forward by a certain number of steps. This can be beneficial for tasks where the sequence of data points is critical as it enables the generation of new data points, similar to the original data but with a unique temporal arrangement.

Flipping, on the other hand, comes in two forms - horizontal and vertical. Horizontal flipping, which reverses the time series sequence, is inappropriate for our study as the temporal order is of utmost importance. Hence, horizontal flipping has been excluded from our data augmentation strategies. Vertical flipping doesn't disrupt the chronological order; rather, it inverts the values, akin to reflecting the time series along the horizontal axis. This strategy can be pertinent in the context of building energy data. For example, chilled water and hot water meter readings are often inversely correlated due to their opposing operational nature; the flipped representation of such time series may provide additional context for model training. Therefore, our study used vertical flipping as a data augmentation method.

Despite the fact that the existence of other data augmentation methods, such as those outlined in relevant work \cite{iwana2021empirical, demir2021data}, this study specifically opted for flipping and shifting as they effectively preserve the inherent structure and temporal sequence of the original data.
By applying these techniques, the initial dataset was expanded fourfold, resulting in approximately three times more data, equivalent to over 5000 time series.

\begin{figure}[!htb]
\begin{center}
\includegraphics[width=0.47\textwidth, trim= 0cm 0cm 0cm 0cm,clip]{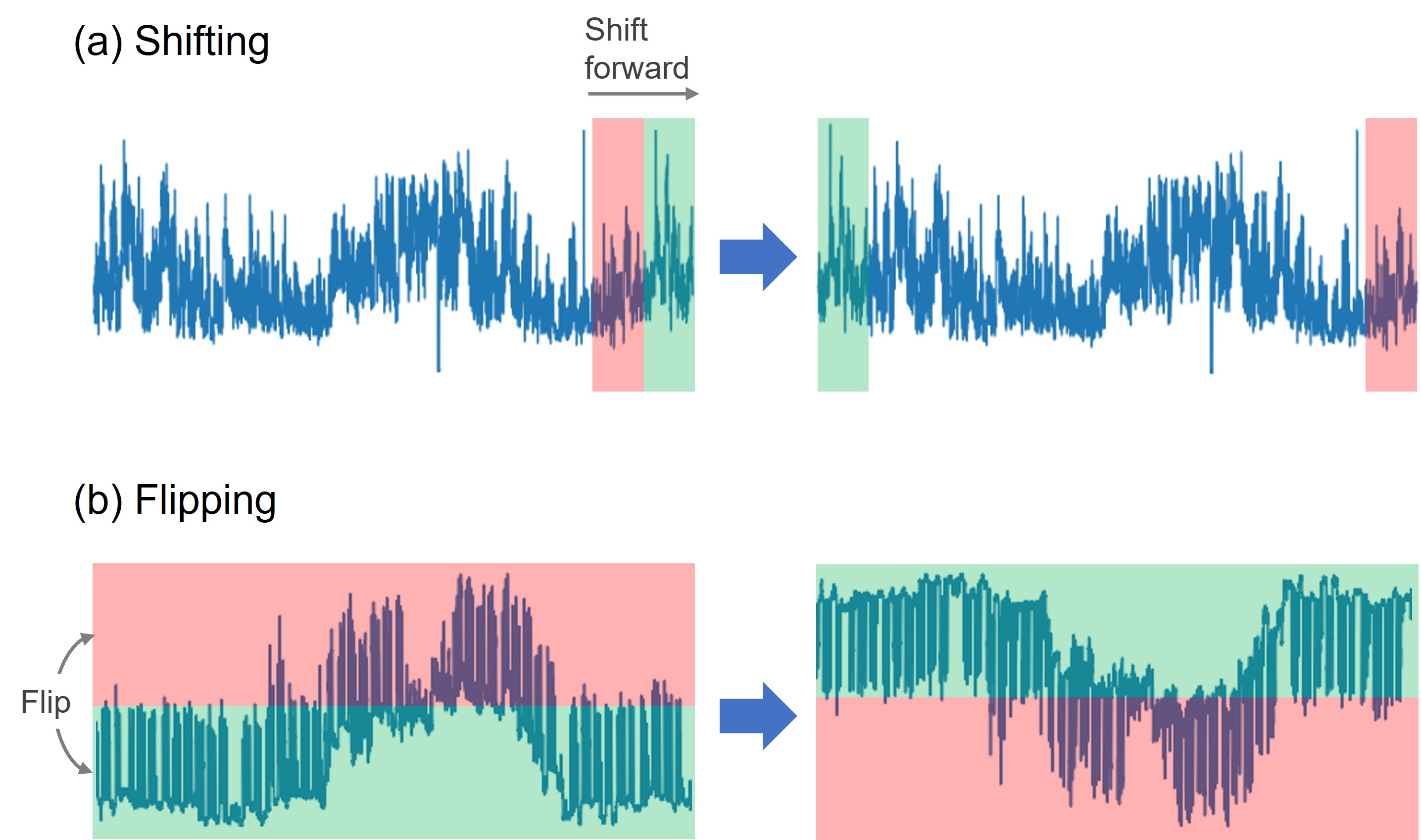}
\caption{Data augmentation: (a) Shifting: shifting time series by a certain number of timesteps (b) Flipping: flipping time series in the vertical direction.}
\label{fig:data_augmentation}
\end{center}
\end{figure}

\subsection{Missing masks}
\begin{figure}[!htb]
\begin{center}
\includegraphics[width=0.32\textwidth, trim= 0cm 0cm 0cm 0cm,clip]{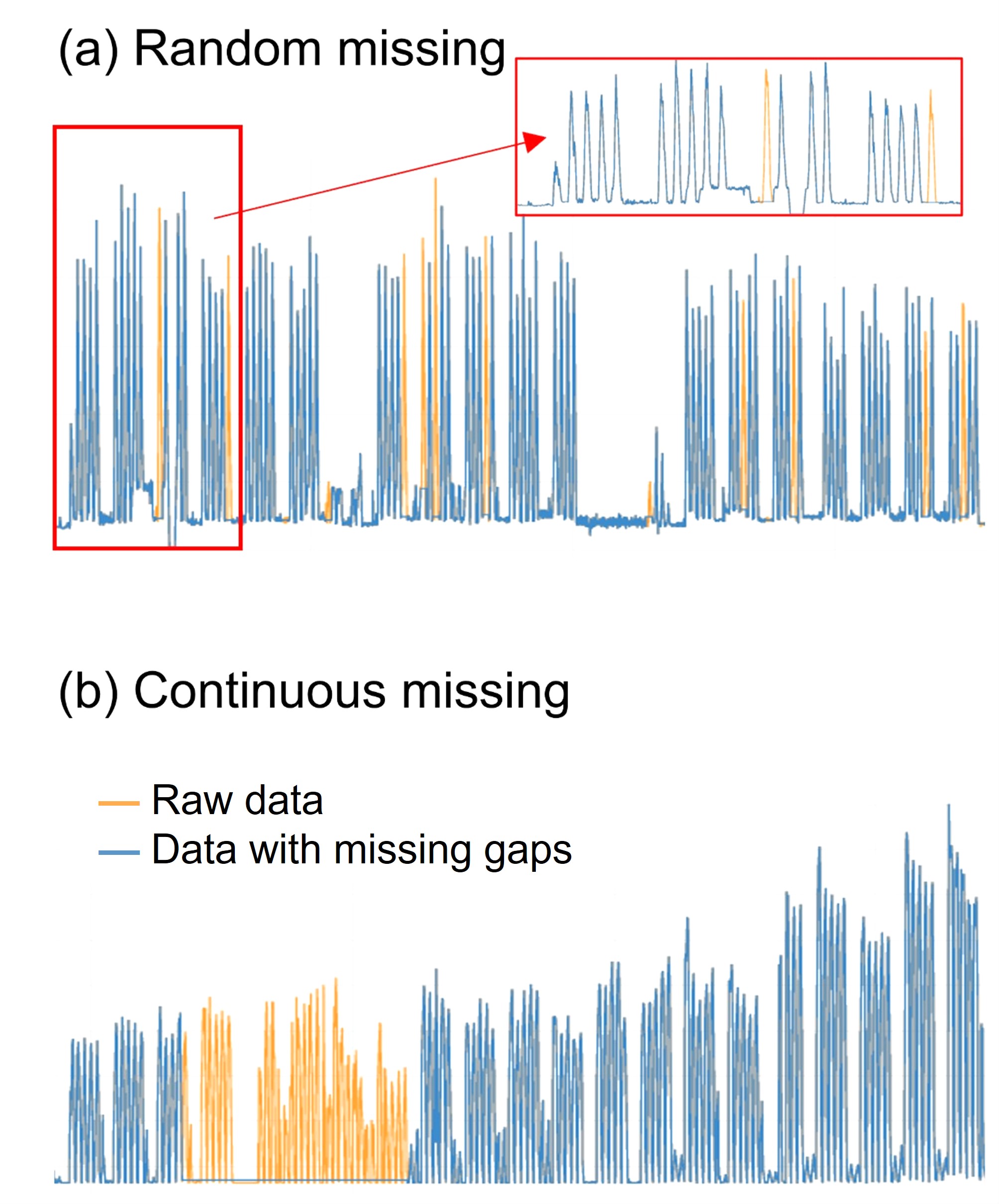}
\caption{Missing type: (a) Random missing: randomly selected days of missing data (b) Continuous missing: consecutive days of missing data}
\label{fig:missing_type}
\end{center}
\end{figure}

Missing masks are synthetic filters used to train imputation models to predict missing gaps in energy data. 
There are two common types of missing gaps in energy data: random missing, resulting from occasional events, and continuous missing, caused by system shutdown or sensor failure (as shown in Figure \ref{fig:missing_type}).
Both types of missing data are included in the imputation tasks of this study and are given a missing rate range of 5\% to 50\%.

\subsubsection{Random missing}
Random missing (also known as short-term missing), which refers to gaps in time series data that are randomly scattered over time, has garnered attention in previous studies. 
The granularity of the missing data, such as hourly or daily intervals, can vary to assess a model's ability to handle different time scales. 
In this study, daily granularity is utilized for the missing gaps. 
Prior research has primarily focused on imputing hourly intervals, and their results indicate that this task is relatively simple \cite{ma2020bi, liguori2021indoor}. 
Hourly missing data can be accurately estimated by nearby values, and even classic statistical methods can achieve satisfactory results \cite{zhang2020pattern, wang2021towards}. 
In contrast, missing entire days at random poses a more significant challenge for an imputation model. 
To evaluate the performance of an imputation model under various missing levels, a missing rate from 5\% to 50\% is employed.

\subsubsection{Continuous missing}
Continuous or consecutive missing (also known as long-term missing) refers to gaps in time series data that occur over a prolonged period, such as missing data that lasts for a week or a month. Imputing this type of missing data can be more challenging as there are fewer nearby data points for reference. Nonetheless, deep learning architectures, particularly those used in computer vision, have successfully addressed this issue by utilizing the data structure to impute the missing values. Similar to the setting of random missing data, a range of 5\% to 50\% continuous missing rate is chosen to evaluate the model's performance.

\subsection{Modeling}
This study explored several modeling approaches for imputing missing energy data, including a weekly persistence model as a na\"ive baseline, CNN-based models, and PConv.
Table \ref{tab:model_table}) provides a summary of the models used and their respective settings. 

\begin{table}[!htb]
\centering
\resizebox{0.47\textwidth}{!}{%
\begin{tabular}{lll}
\hline
\textbf{Model} & \textbf{Hyperparameters} & \textbf{\begin{tabular}[c]{@{}l@{}}Dimensions of data\\ (samples, dim1, dim2)\end{tabular}} \\ \hline
\begin{tabular}[c]{@{}l@{}}Weekly \\ persistence \end{tabular} & None & (1479, 8736, 1) \\
 &  &  \\
1D-CNN & \begin{tabular}[c]{@{}l@{}}encoder = 3   layers; decoder = 3 layers; \\ fully-connected layer at bottleneck\end{tabular} & (1479, 8736, 1) \\
 &  &  \\
2D-CNN & \begin{tabular}[c]{@{}l@{}}encoder = 2   layers; decoder = 2 layers; \\ fully-connected layer at bottleneck\end{tabular} & (1479, 168, 52) \\
 &  &  \\
PConv & encoder = 4 layers; decoder = 4 layers & (1479, 168, 52) \\ \hline
\end{tabular}
}
\caption{Overview of models and na\"ive baseline regarding hyperparameters and dimensions of data.}
\label{tab:model_table}
\end{table}

\subsubsection{Weekly persistence model}
The weekly persistence model, a na\"ive method, was implemented in this study as a baseline for comparison with other imputation approaches \cite{de2016practical, sauter2018load}. 
Unlike sophisticated algorithms like deep neural networks or regression methods, the weekly persistence model operates on the assumption that energy consumption patterns persist over time. 
Specifically, it predicts the energy consumption at a given time in the current week by referencing the energy consumption at the same time in the previous week.
For instance, the consumption value at 8:00 AM on a Wednesday is used to predict the energy usage for the subsequent Wednesday at 8:00 AM.
Despite its simplicity, the weekly persistence model has been effectively used in energy modeling as a benchmark and has, in some scenarios, outperformed classical statistical methods \cite{weber2021data}. Its widespread use and demonstrated effectiveness make it a valuable comparison point for evaluating the performance of more advanced imputation methods.

\subsubsection{Convolutional Neural Networks (CNNs)}
Convolutional Neural Networks (CNNs) are a type of neural network that is particularly well-suited for processing data with a spatial structure, such as images or time series.
Compared to Fully-Connected Neural Networks (FCNNs), CNNs differ in that they learn local patterns in their input feature space, while FCNNs learn global patterns. 
This means that after a CNN has successfully learned a pattern in one location, it can recognize it in another location with minimal effort.
In contrast, an FCNN would have to learn the pattern again if it appeared in a different location. 
As a result, CNNs are more efficient when dealing with multidimensional data, as they require fewer training samples to learn representations with high generalization power. 
Thus, we used two CNN frameworks for data imputation: a one-dimensional CNN (1D-CNN) for time series data and a two-dimensional CNN (2D-CNN) for reshaped data.
To fit energy data to a 2D-CNN, the data is reshaped with the time of the week and week number serving as the two dimensions. 
For example, suppose the energy data from a power meter consists of hourly readings over a year. In that case, it is reshaped into a matrix where rows represent the time of the week (e.g., hour 1, hour 2, \dots, hour 168) and columns represent the week number (e.g., week 1, week 2, \dots, week 52). 
The reshaped data with the dimensions ($168\times 52$) is then fed into the 2D-CNN for processing and imputation of missing values.

The CNNs in this study were implemented using an autoencoder architecture, consisting of two components: an encoder and a decoder. 
The encoder converts the input data into a compressed representation with reduced dimensions. 
This compression process could capture and extract the essential features of the input data in a more compact form, enabling efficient representation learning.
Subsequently, the decoder component receives the compressed representation generated by the encoder and reconstructs the original data from this compressed representation. 
The decoder's role is to transform the compressed representation back into its original format, facilitating the recovery of the complete information. 
During the training process, the autoencoder aims to minimize the difference between the original data and the reconstructed data, encouraging the model to learn an effective compression and reconstruction process.
In the context of imputing missing values in energy data, the autoencoder is provided with data containing synthetic missing gaps, aiming to predict the original complete data as its output. 

\subsubsection{Partial Convolution (PConv)}
\begin{figure*}[!htb]
\begin{center}
\includegraphics[width=0.7\textwidth, trim= 0cm 0cm 0cm 0cm,clip]{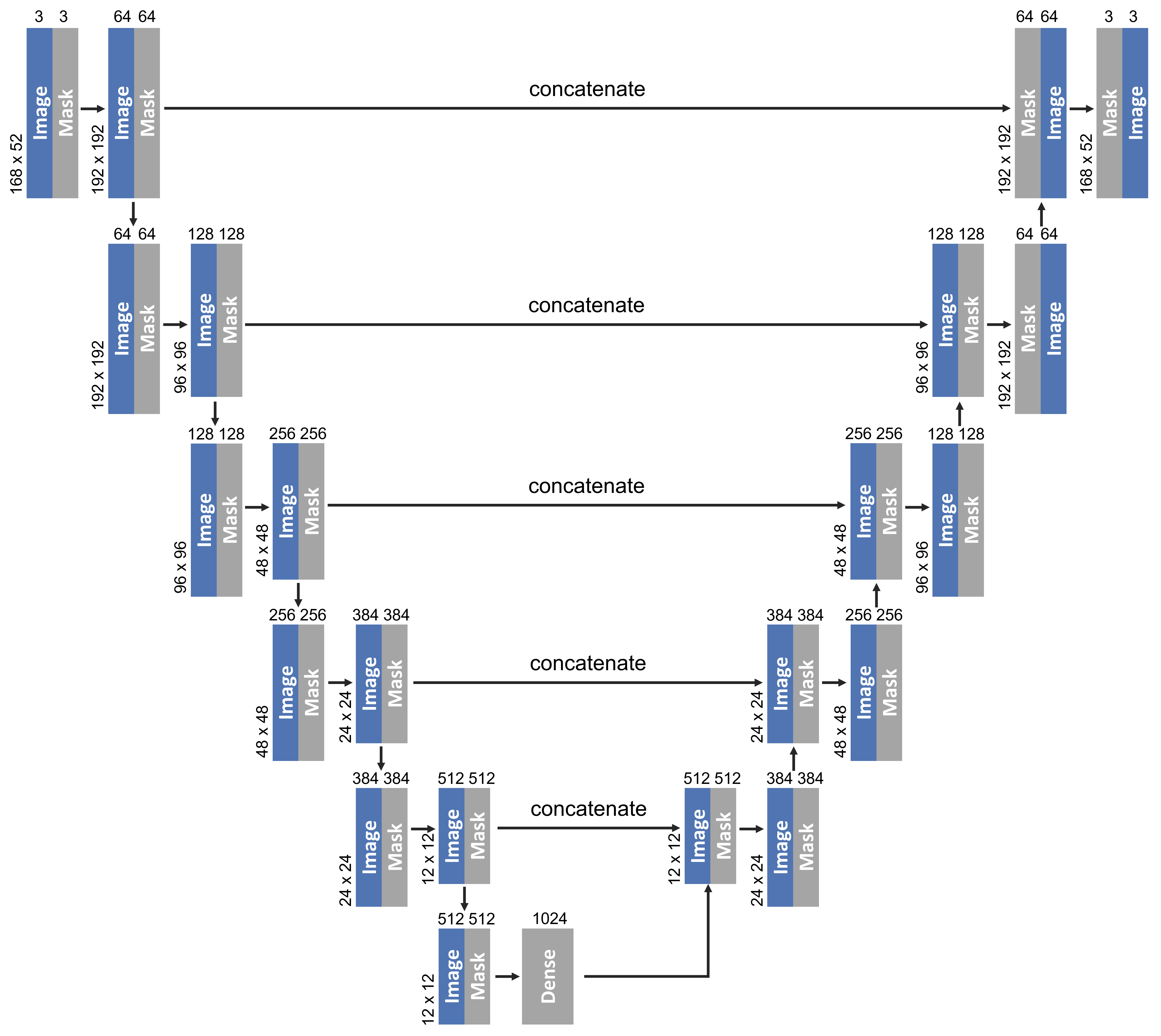}
\caption{U-net architecture of Pconv for imputing two-dimensional energy data.}
\label{fig:pconv_diagram}
\end{center}
\end{figure*}

In addition to utilizing traditional CNN models for imputing missing energy data, an advanced deep learning framework called Partial Convolution (PConv) was also tested to explore the potential of a computer-vision inpainting technique in the energy domain. 
One of the main advantages of PConv is its automatic transmission mask mechanism, which allows the model to identify the areas that need to be repaired and use this information to achieve state-of-the-art results. 
The PConv model was implemented using the U-Net architecture, modified to use partial convolutions instead of regular convolutions. 
The final architecture is shown in Figure \ref{fig:pconv_diagram}.
Notably, PConv excels in handling irregularly shaped missing areas, as opposed to traditional image repair techniques that typically only deal with regular gaps (e.g., rectangular and circle shapes). 
This makes PConv particularly robust and well-suited for a wide range of missing data scenarios.

While there are other advanced deep learning frameworks for image inpainting, such as Generative Adversarial Networks (GANs) and diffusion models, they can be difficult to implement and train. 
GANs, in particular, can be challenging to train due to the need to balance the generator and discriminator and the instability of the minimax loss function \cite{arjovsky2017wasserstein, arjovsky2017towards}. 
In contrast, PConv does not require a discriminator to be trained and has a more simplified model architecture, making it more suitable for small datasets like the building energy dataset used in this study. 
Regarding the input data for training PConv, the energy time series data was preprocessed into a two-dimensional matrix, identical to the input format for 2D-CNN models.

\subsubsection{Model hyperparameters setting and training process}
All of the models compared in this study, except for the weekly persistence model, are based on neural networks. 
The performance of these models is significantly influenced by their hyperparameters, such as the number of layers and channels. 
For the 1D- and 2D-CNN models, the hyperparameters were based on the autoencoder example provided on the Keras official website \footnote{\url{https://blog.keras.io/building-autoencoders-in-keras.html}}.
Keras is a popular high-level API for deep learning, built on top of TensorFlow, one of the most widely used deep learning frameworks.
To expand the model's capacity for learning, a fully-connected layer is added at the bottleneck, which enables the model to learn complex relationships between the input and output data by allowing information to flow freely between all neurons in the layer. 

The hyperparameters for the PConv model were adopted from the original paper by Liu et al. \cite{liu2018image}. However, due to memory constraints in the Google Colab platform\footnote{https://colab.research.google.com/} used in this study, the image size was adjusted to 192$\times$192 pixels from the original 256$\times$256 pixels. It's worth noting, however, that the resolution of 192$\times$192 pixels is sufficiently large to accommodate the original dimensions of 168$\times$52, thus ensuring that no significant loss of data or features occurs in the resizing process. The dimensions of the reshaped time series were resized from 168$\times$52 to 192$\times$192 using interpolation. To maintain data integrity, the dimensions were resized to 192$\times$192 using interpolation, a method that retains the key structure and trends of the data. An overview of the chosen hyperparameters for the models can be found in Table \ref{tab:model_table}.

\begin{figure}[!htb]
\begin{center}
\includegraphics[width=0.4\textwidth, trim= 0cm 0cm 0cm 0cm,clip]{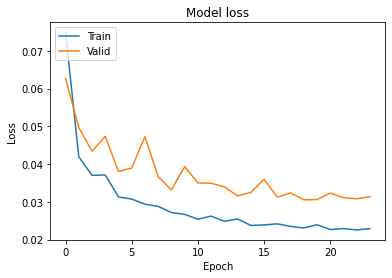}
\caption{Training and validation loss over epochs for the PConv model.}
\label{fig:pconv_loss}
\end{center}
\end{figure}

In this study, we employed several strategies to enhance model training and ensure the robustness of the results. 
To facilitate model training, both random and continuous synthetic missing masks with varying rates were randomly applied to the training dataset, along with additional irregular missing masks, which were suggested in the original paper\cite{liu2018image}.
Furthermore, an early stopping mechanism was implemented in the PConv and CNN models to monitor the convergence of the model training process. This mechanism observes the model's performance on a validation set comprising 20\% of the total dataset and stops the training process when the validation loss fails to improve over five consecutive epochs. The training process is allowed to continue for a total of 50 epochs unless halted earlier by the early stopping mechanism. Figure \ref{fig:pconv_loss} provides an example of the loss over epochs for the PConv model, demonstrating the convergence of training and validation loss and the triggering of the early stop when the patience of 5 epochs was reached due to the lack of improvement in the validation loss.

\subsubsection{Evalaution metrics}
To assess the model's performance for imputing missing gaps, we employed the Mean Squared Error (MSE) and R-squared metric, both of which are widely accepted measures for assessing model performance in time series prediction tasks, particularly in energy prediction. 
MSE is a commonly used measure of prediction error that calculates the average squared difference between the predicted and actual values. 
This metric is used to quantify the prediction error between the predicted results and the ground truth on synthetic missing energy data. 
R-squared, also known as the coefficient of determination, quantifies the extent to which the trend of the predicted values aligns with the actual values. 
In energy time series data, it is crucial to predict trends and patterns accurately, and R-squared can help determine whether the model can capture these trends and patterns.
Though Normalized Root Mean Squared Error (NRMSE) is another commonly used metric in this field, we opted not to incorporate it into our evaluation. This decision was due to the pre-processing step where all energy meter data were normalized via a max-min normalization technique prior to the application of the inpainting algorithm, thereby limiting the added value that the NRMSE metric could offer in this context.

\subsection{Experiment design}
Table \ref{tab:exp_design} presents the settings for the experiment, which are based on the data preprocessing, missing masks, and modeling techniques described in previous sections. 
This experimental design will allow us to evaluate the model's generalization performance and assess its ability to handle missing data of various degrees under different scenarios.

\begin{table}[!htb]
\centering
\resizebox{0.35\textwidth}{!}{%
\begin{tabular}{ll}
\hline
 & \textbf{Setting of experiments} \\ \hline
\begin{tabular}[c]{@{}l@{}}Models and data \\ dimensions\end{tabular} & \begin{tabular}[c]{@{}l@{}}- Weekly persistence model (1D)\\ - 1D-CNN (1D)\\ - 2D-CNN (2D)\\ - PConv (2D)\end{tabular} \\
 &  \\
Missing masks & \begin{tabular}[c]{@{}l@{}}Random and continuous days with \\ missing rates between 5 and 50\%\end{tabular} \\
 &  \\
Cross-validation & Five folds split by site ID \\
 &  \\
Data augmentation & Shifting and flipping time series \\ \hline
\end{tabular}
}
\caption{Overview of modeling and experiment settings.}
\label{tab:exp_design}
\end{table}

\section{Results}
\label{sec:results}
In the result section, we will discuss and interpret our findings pertaining to the prediction error values between the models and explore how the type of meter (e.g., electricity and hot water meters) affects the predictability of missing values. 
To further demonstrate the effectiveness of image imputation in energy data, examples in the form of trend plots (1D) and heatmaps (2D) from meters will also be presented to show how context can assist in the data imputation process for each model.

\subsection{Random and continuous missing}
\begin{figure*}[!htb]
\centering

\begin{subfigure}{0.8\textwidth}
    \includegraphics[width=\textwidth]{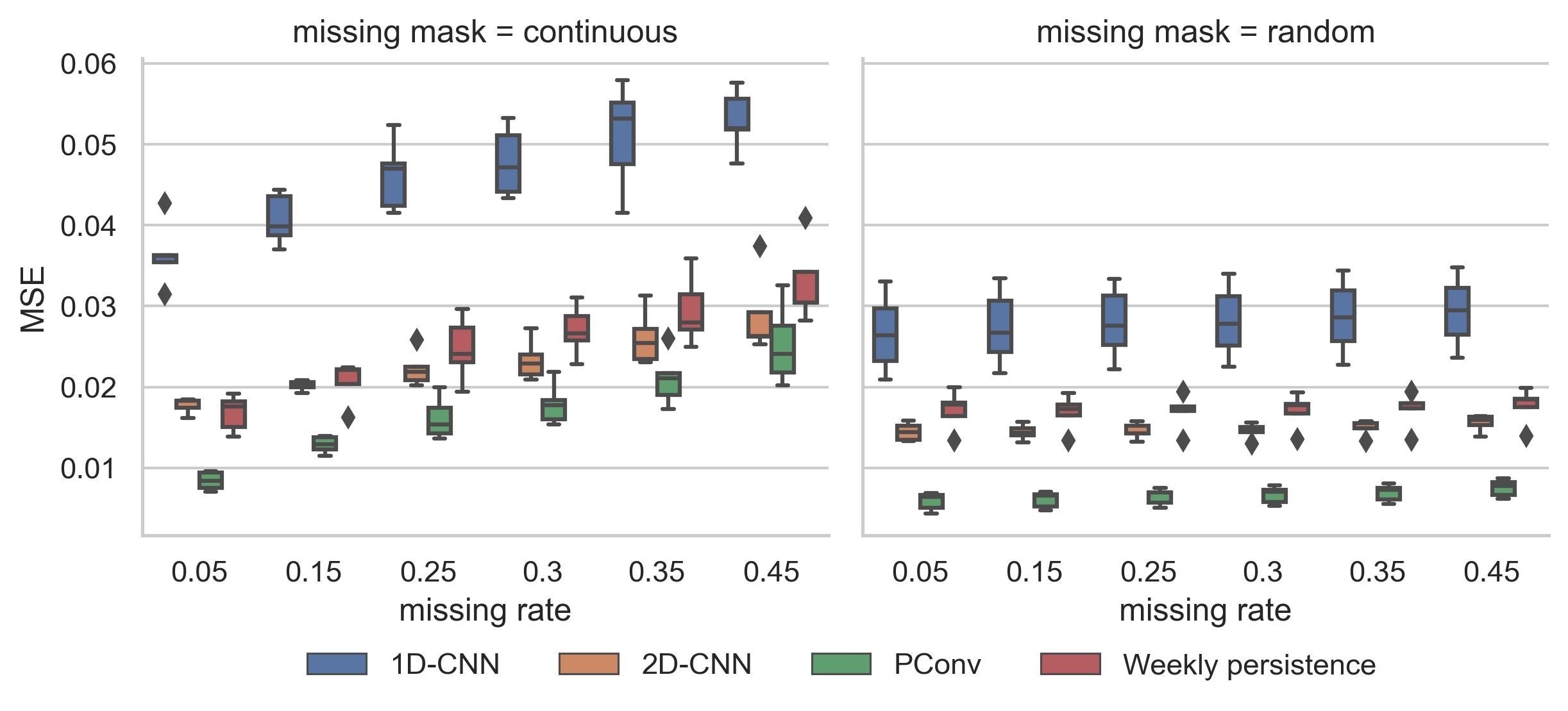}
    %\caption{Firts subfigure.}
    %\label{fig:first}
\end{subfigure}

\vspace{-0.6cm}

\begin{subfigure}{0.8\textwidth}
    \includegraphics[width=\textwidth]{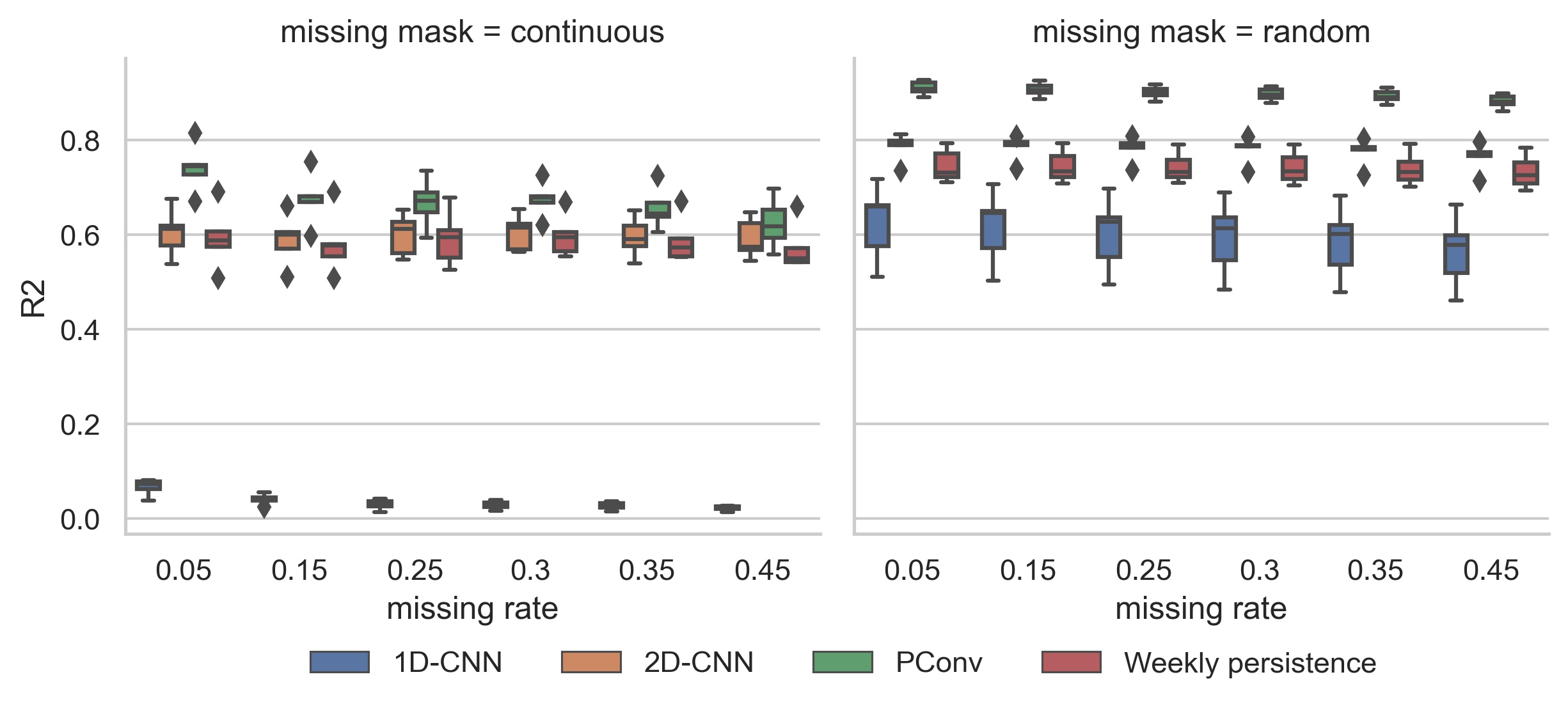}
    %\caption{Second subfigure.}
    %\label{fig:second}
\end{subfigure}

\caption{Comparison between models in different imputation tasks with varying missing rates and types.}
\label{fig:mse_missingRate}
\end{figure*}

\begin{figure*}[!htb]
\begin{center}
\includegraphics[width=0.8\textwidth, trim= 0cm 0cm 0cm 0cm,clip]{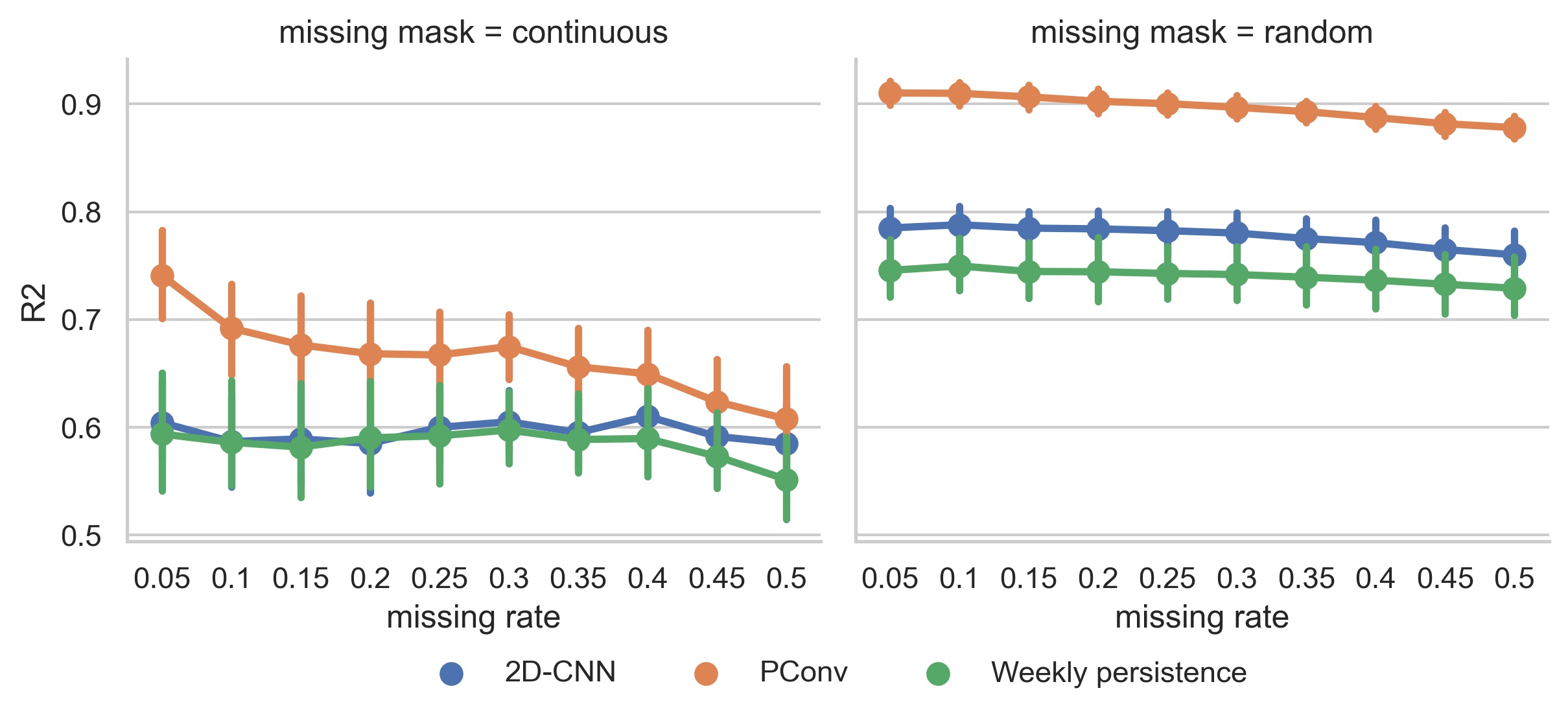}
\caption{Zooming in on R\textsuperscript{2} comparison to observe the decline of imputation performance as the missing rate increases}
\label{fig:R2_zoom_in}
\end{center}
\end{figure*}

To investigate whether the extra dimension from reshaped time series impacts the imputation of missing values, the prediction errors of the models under different missing categories are comprehensively compared in Figure \ref{fig:mse_missingRate}.
Overall, continuous missing data poses a higher challenge for imputation in comparison to random missing data. This is due to the fact that imputing random missing gaps can leverage more context from neighboring values. 
When comparing different models' performance in imputing continuous missing data, we observe that the 1D-CNN model significantly underperforms in comparison to two-dimensional models (i.e., 2D-CNN and PConv) and the weekly persistence method. 
In contrast, PConv, as an image-inpainting framework, demonstrates superior results with an MSE of 0.017, significantly lower than other models by 27 to 61\%.
As for data imputation results for random missing on the right boxplots, each model performs much better than continuous missing data, and PConv overwhelmingly outperforms other models, with an average MSE 50\% lower than the next best model (i.e., 2D-CNN).
The results show that two-dimensional methods with reshaped energy data considerably outperform the one-dimensional and na"ive methods. 
Especially, PConv has better performance than 2D-CNN, showcasing the efficacy of image-based techniques in missing energy data imputation.
However, it is essential to note that a higher missing rate for continuous data could significantly impact the performance of the models. For instance, when the missing rate for continuous data rises from 5\% to 10\%, the average R\textsuperscript{2} of PConv significantly decreases from 0.75 to less than 0.7 (as shown in Figure \ref{fig:R2_zoom_in}). This highlights the fundamental difficulty of imputing data when the missing rate exceeds 10\% of the annual data, roughly equivalent to a two-week absence of data.

\subsection{Breakdown by meter types}
\begin{figure*}[!htb]
\centering

\begin{subfigure}{0.75\textwidth}
    \includegraphics[width=\textwidth]{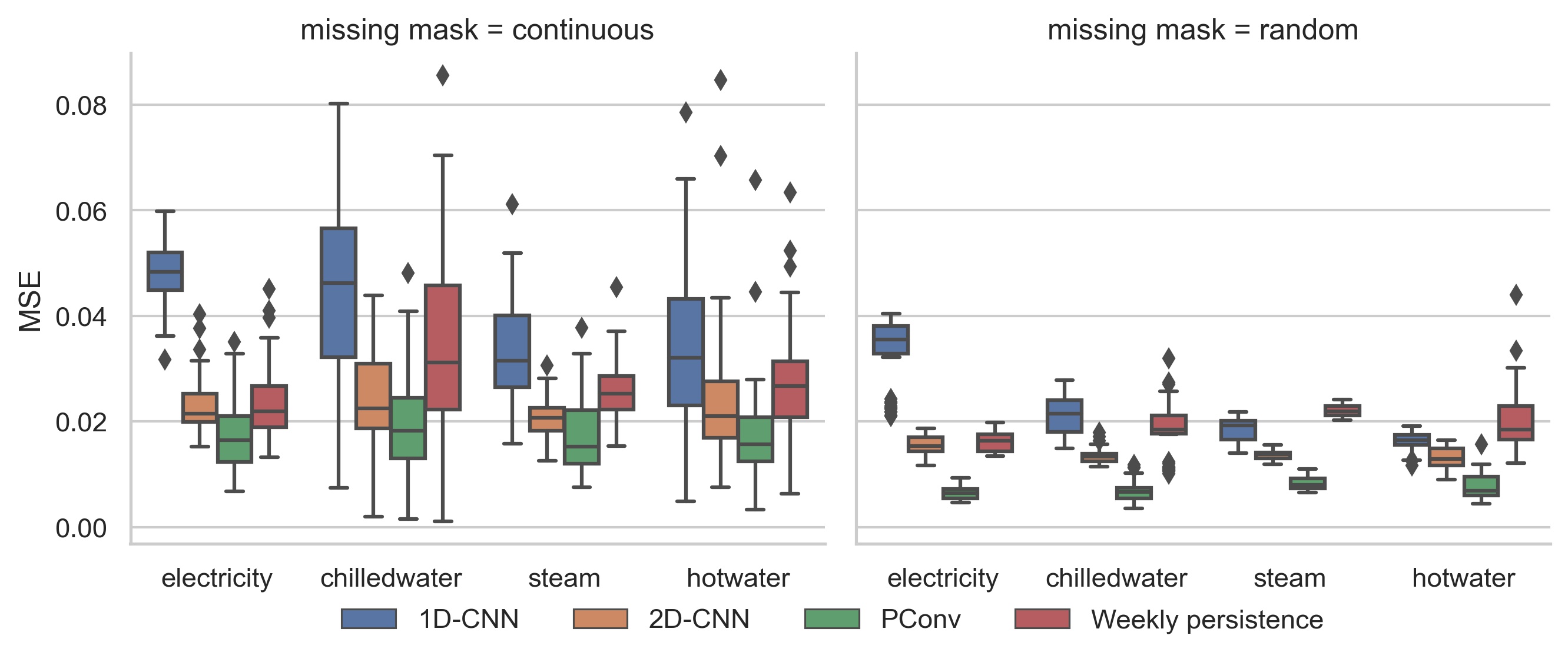}
    %\caption{Firts subfigure.}
    %\label{fig:first}
\end{subfigure}

\vspace{-0.5cm}

\begin{subfigure}{0.75\textwidth}
    \includegraphics[width=\textwidth]{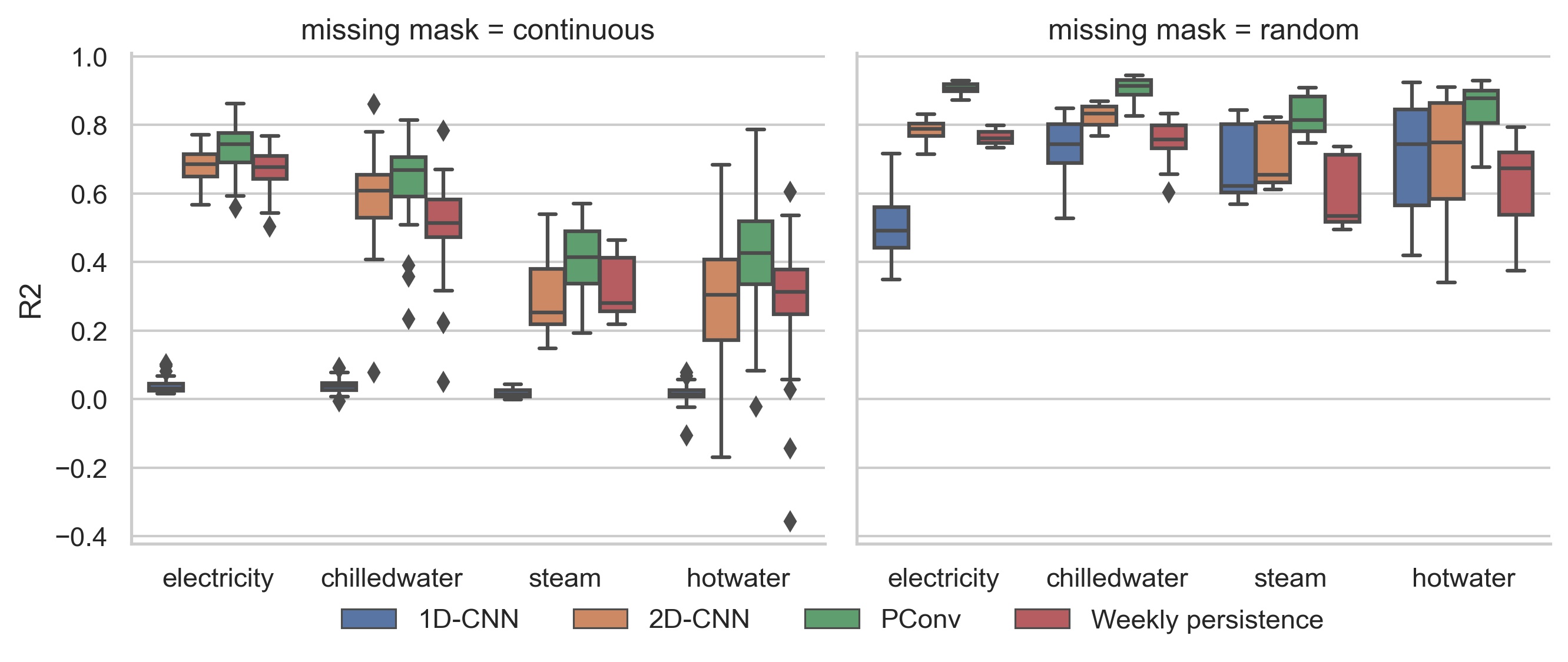}
    %\caption{Second subfigure.}
    %\label{fig:second}
\end{subfigure}

\caption{Breakdown of model performance according to meter types.}
\label{fig:mse_meterType}
\end{figure*}

The effect of different meter types on imputation results is further explored and shown in Figure \ref{fig:mse_meterType}. 
Generally, PConv offers the most accurate predictions, followed by 2D-CNN, the weekly persistence method, and finally, 1D-CNN. 
Interestingly, the weekly persistence method performs comparably to the 2D-CNN in predicting electricity meters.
 This could be attributed to the more regular patterns present in electricity meter data, which allows the weekly persistence method to predict values accurately by mimicking values from the previous week. 
In contrast, meters highly influenced by weather conditions, such as chiller water, steam, and hot water meters, present greater prediction difficulties for the weekly persistence method due to their inherent irregularity. 
Despite PConv's superior performance across other models, it still struggles with predicting the continuous missing data for weather-dependent meters. 
For example, as demonstrated by the bottom-left boxplots, PConv's imputation on steam and hot water meters cannot even reach an R squared of 0.6 for continuous missing data. 
This underscores the challenge that stems from not only the continuity of missing data and its rate but also the dependencies of the meters on weather conditions. 

%\begin{figure}[!htb]
%\begin{center}
%\includegraphics[width=1.0\textwidth, trim= 0cm 0cm 0cm 0cm,clip]{Figures/mse_meterType.jpg}
%\caption{Breakdown of model performance according to meter types.}
%\label{fig:mse_meterType}
%\end{center}
%\end{figure}

\newpage
\subsection{Zoom-in of imputation results from visualized examples}
\begin{figure*}[!htb]
\centering

\begin{subfigure}{0.8\textwidth}
    \caption{10\% Random missing (electricity meter).}
    \includegraphics[width=\textwidth]{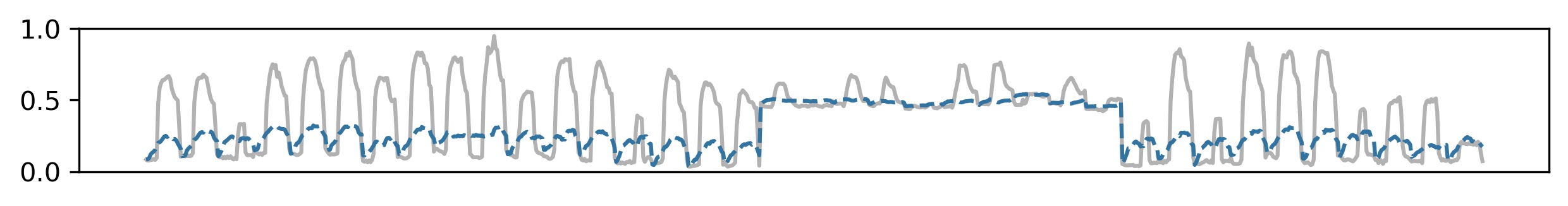}
    \label{fig:ts_plots_c}
\end{subfigure}

\vspace{-0.7cm}

\begin{subfigure}{0.8\textwidth}
    %\caption{10\% Continuous missing (electricity meter).}
    \includegraphics[width=\textwidth]{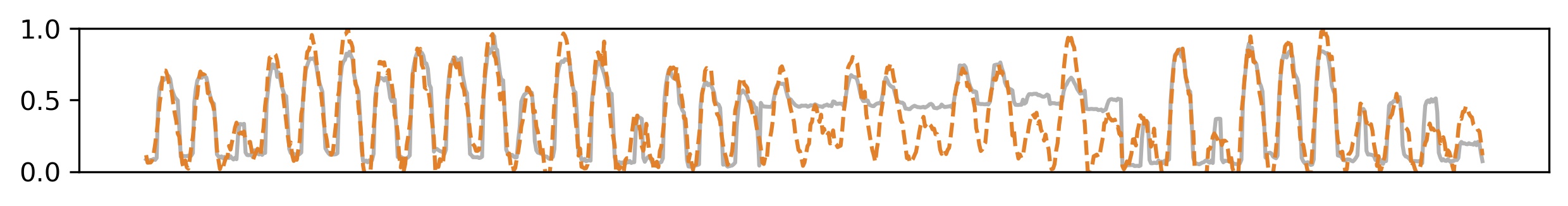}
    %\label{fig:first}
\end{subfigure}

\begin{subfigure}{0.8\textwidth}
    %\caption{10\% Continuous missing (electricity meter).}
    \includegraphics[width=\textwidth]{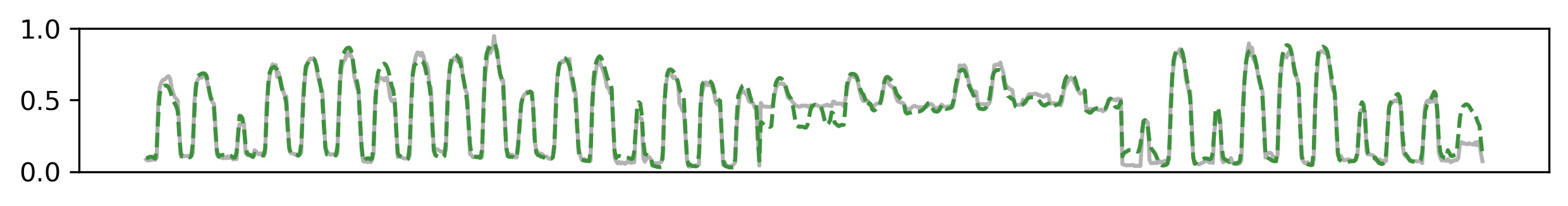}
    %\label{fig:first}
\end{subfigure}

\begin{subfigure}{0.8\textwidth}
    %\caption{10\% Continuous missing (electricity meter).}
    \includegraphics[width=\textwidth]{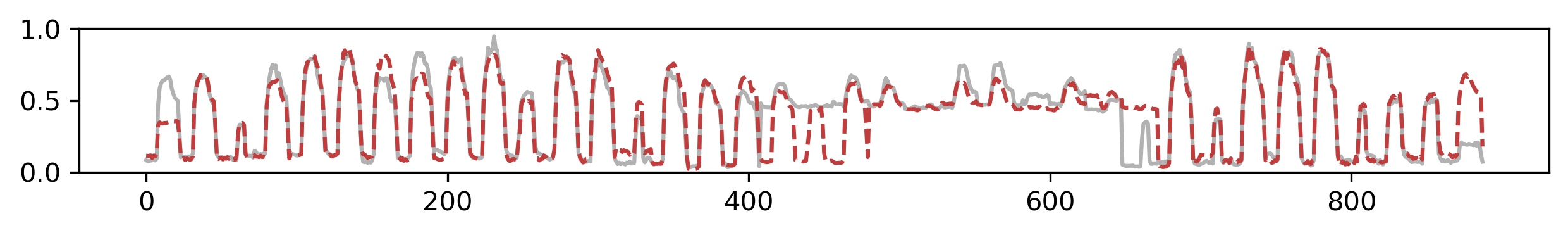}
    %\label{fig:first}
\end{subfigure}

\begin{subfigure}{0.8\textwidth}
    \caption{10\% Random missing (hot water meter).}
    \includegraphics[width=\textwidth]{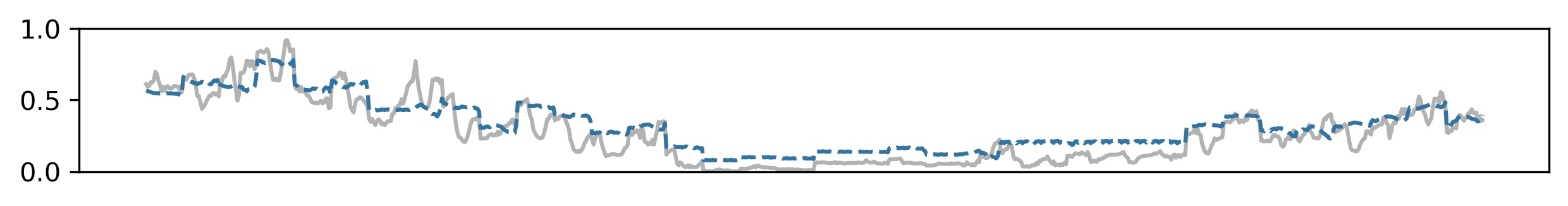}
    \label{fig:ts_plots_d}
\end{subfigure}

\vspace{-0.7cm}

\begin{subfigure}{0.8\textwidth}
    %\caption{10\% Continuous missing (hot water meter).}
    \includegraphics[width=\textwidth]{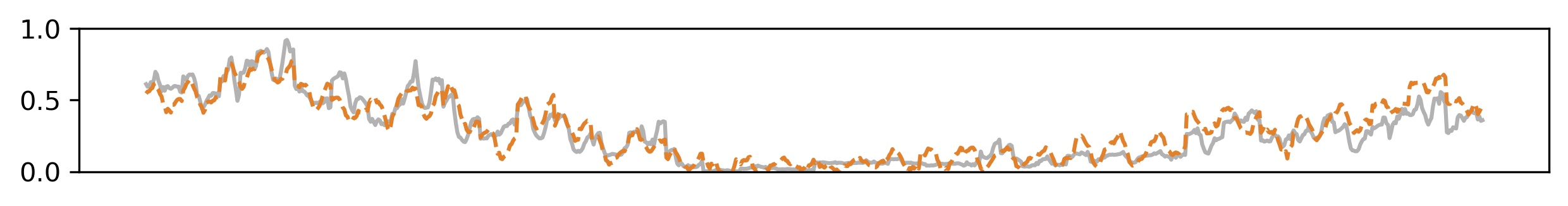}
    %\label{fig:second}
\end{subfigure}

\begin{subfigure}{0.8\textwidth}
    %\caption{10\% Continuous missing (hot water meter).}
    \includegraphics[width=\textwidth]{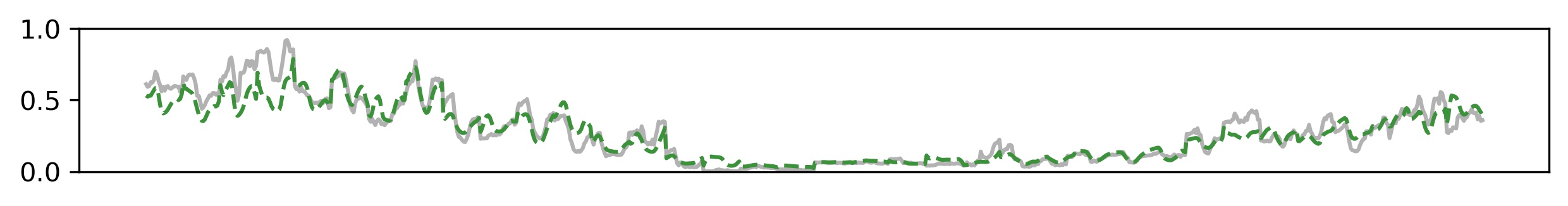}
    %\label{fig:second}
\end{subfigure}

\begin{subfigure}{0.8\textwidth}
    %\caption{10\% Continuous missing (hot water meter).}
    \includegraphics[width=\textwidth]{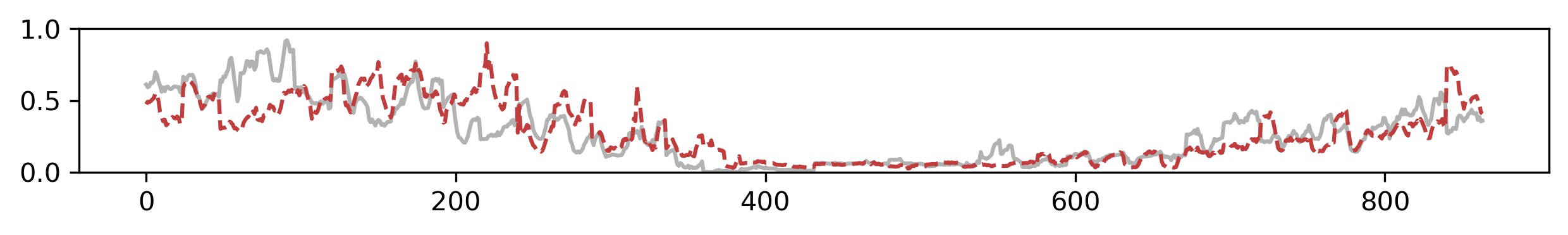}
    %\label{fig:second}
\end{subfigure}

\begin{subfigure}{0.6\textwidth}
    %\caption{10\% Continuous missing (electricity meter).}
    \includegraphics[width=\textwidth]{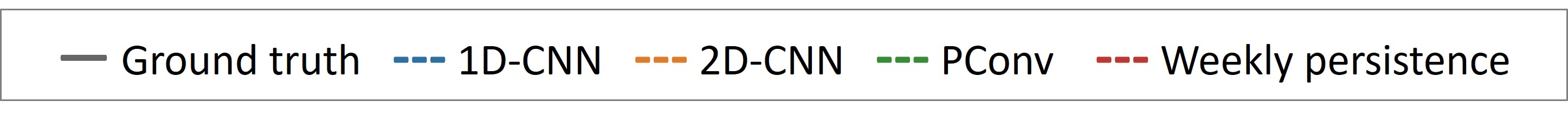}
    %\label{fig:first}
\end{subfigure}

\caption{Time series plots of power meters for comparing imputation results from different methods for 10\% random missing data.}
\label{fig:ts_plots_random}
\end{figure*}

\begin{figure*}[!htb]
\centering

\begin{subfigure}{0.8\textwidth}
    \caption{10\% Continuous missing (electricity meter).}
    \includegraphics[width=\textwidth]{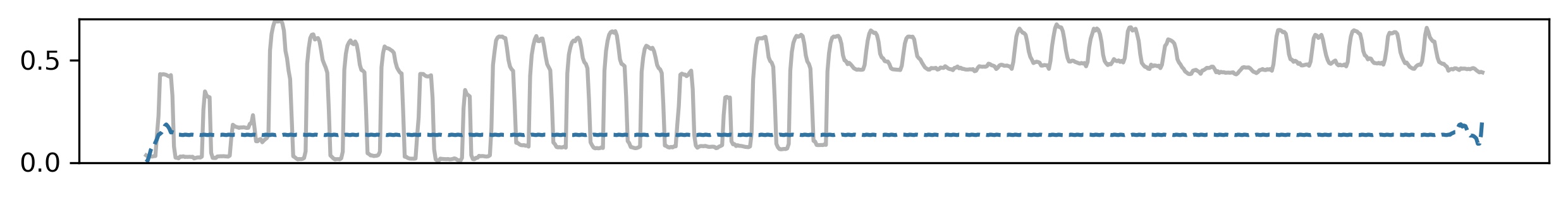}
    \label{fig:ts_plots_a}
\end{subfigure}

\vspace{-0.7cm}

\begin{subfigure}{0.8\textwidth}
    %\caption{10\% Continuous missing (electricity meter).}
    \includegraphics[width=\textwidth]{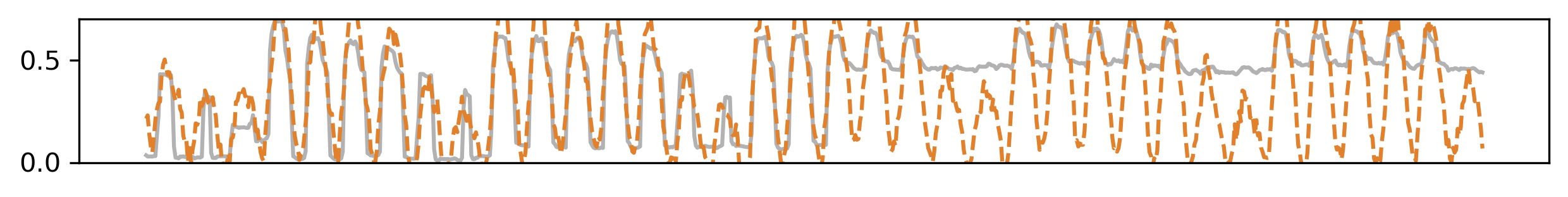}
    %\label{fig:first}
\end{subfigure}

\begin{subfigure}{0.8\textwidth}
    %\caption{10\% Continuous missing (electricity meter).}
    \includegraphics[width=\textwidth]{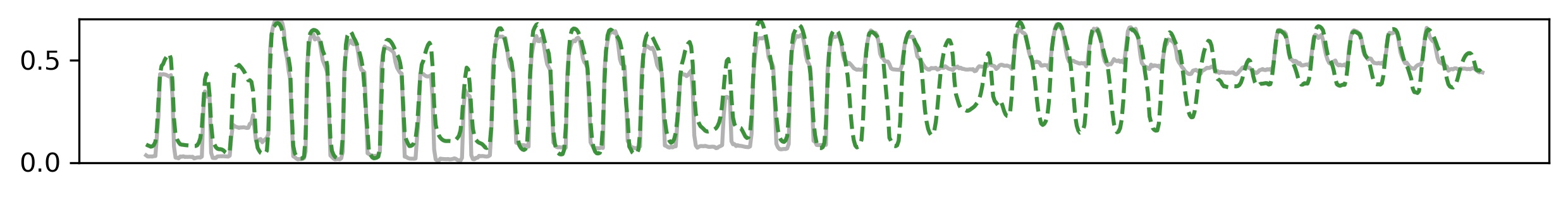}
    %\label{fig:first}
\end{subfigure}

\begin{subfigure}{0.8\textwidth}
    %\caption{10\% Continuous missing (electricity meter).}
    \includegraphics[width=\textwidth]{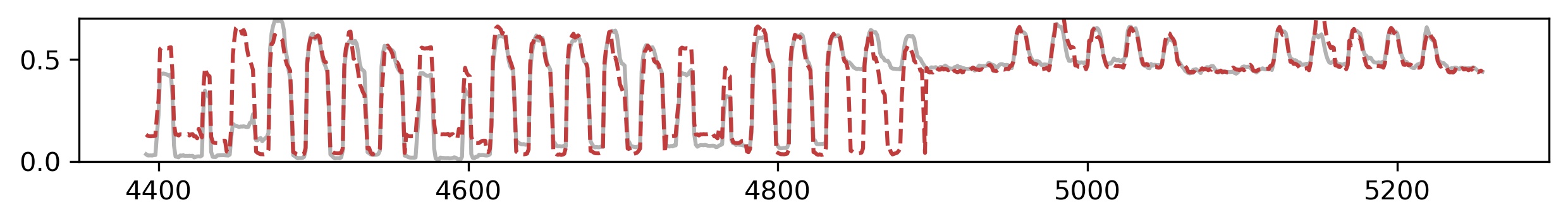}
    %\label{fig:first}
\end{subfigure}

\begin{subfigure}{0.8\textwidth}
    \caption{10\% Continuous missing (hot water meter).}
    \includegraphics[width=\textwidth]{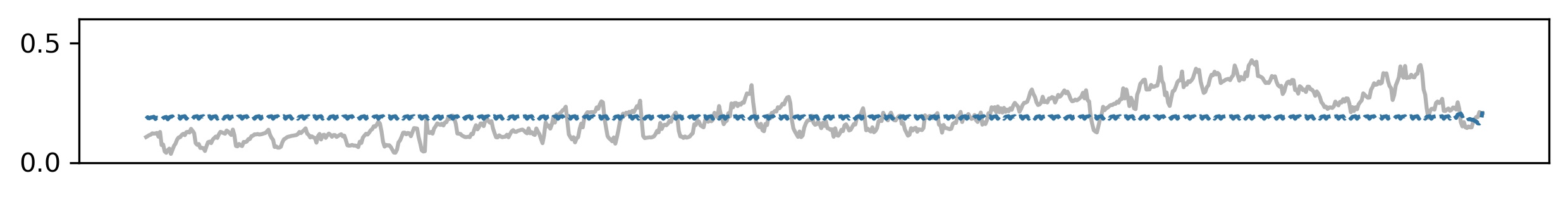}
    \label{fig:ts_plots_b}
\end{subfigure}

\vspace{-0.7cm}

\begin{subfigure}{0.8\textwidth}
    %\caption{10\% Continuous missing (hot water meter).}
    \includegraphics[width=\textwidth]{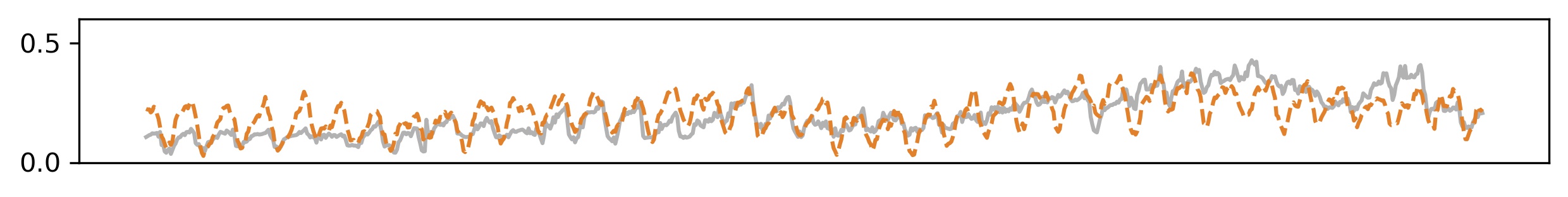}
    %\label{fig:second}
\end{subfigure}

\begin{subfigure}{0.8\textwidth}
    %\caption{10\% Continuous missing (hot water meter).}
    \includegraphics[width=\textwidth]{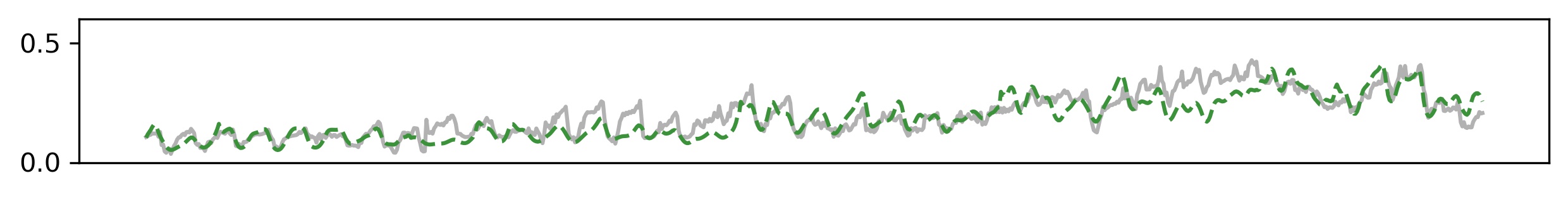}
    %\label{fig:second}
\end{subfigure}

\begin{subfigure}{0.8\textwidth}
    %\caption{10\% Continuous missing (hot water meter).}
    \includegraphics[width=\textwidth]{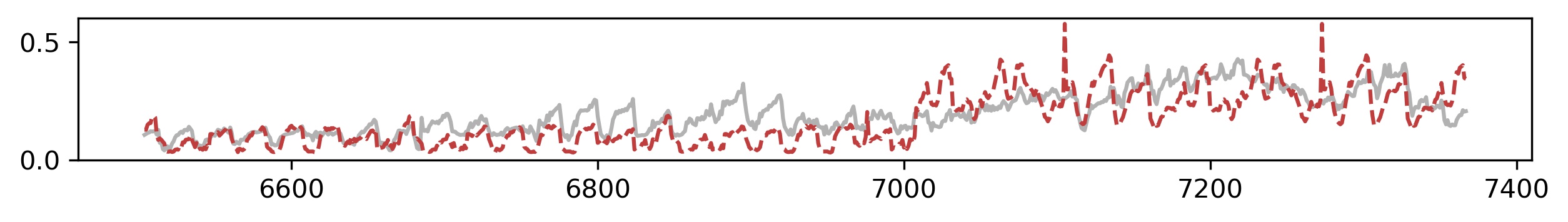}
    %\label{fig:second}
\end{subfigure}

\begin{subfigure}{0.6\textwidth}
    %\caption{10\% Continuous missing (electricity meter).}
    \includegraphics[width=\textwidth]{Figures/TSplot_legend.jpg}
    %\label{fig:first}
\end{subfigure}

\caption{Time series plots of power meters for comparing imputation results from different methods for 10\% continuous missing data.}
\label{fig:ts_plots_continous}
\end{figure*}

\begin{figure*}[!htb]
\centering

\begin{subfigure}{0.8\textwidth}
    \caption{10\% Continuous missing (electricity meter).}
    \includegraphics[width=\textwidth]{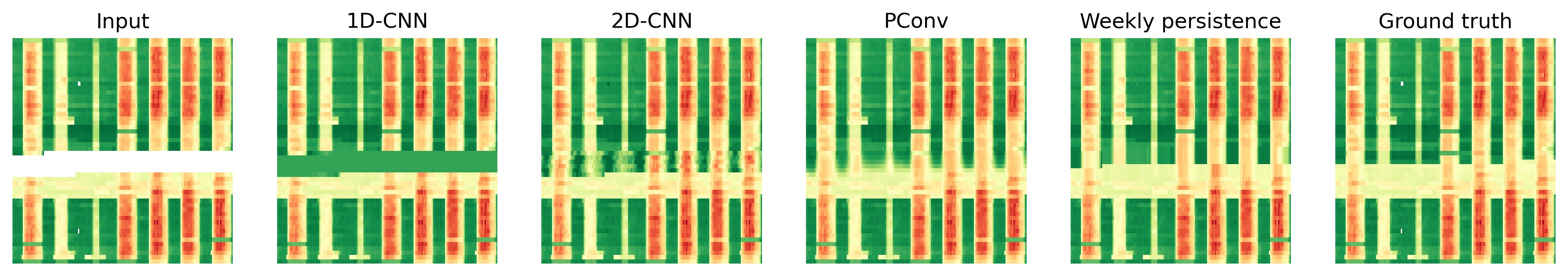}
    %\label{fig:first}
\end{subfigure}

\begin{subfigure}{0.8\textwidth}
    \caption{10\% Continuous missing (hot water meter).}
    \includegraphics[width=\textwidth]{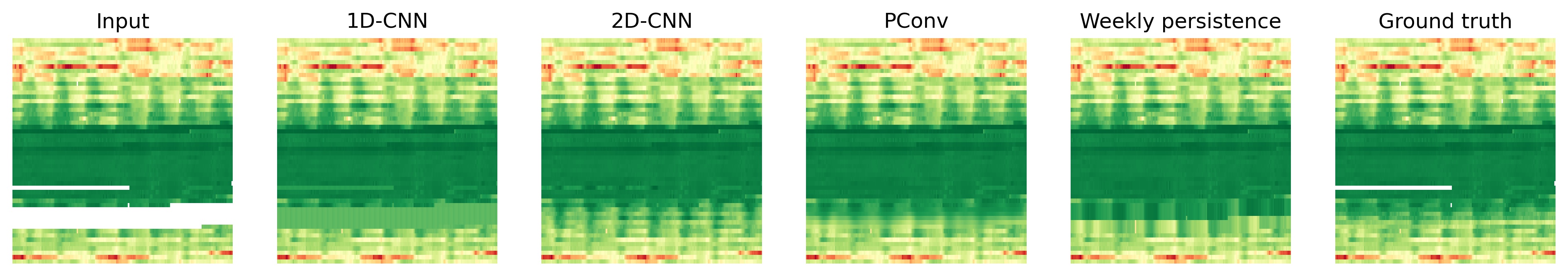}
    %\label{fig:second}
\end{subfigure}

\begin{subfigure}{0.8\textwidth}
    \caption{10\% Random missing (electricity meter)}
    \includegraphics[width=\textwidth]{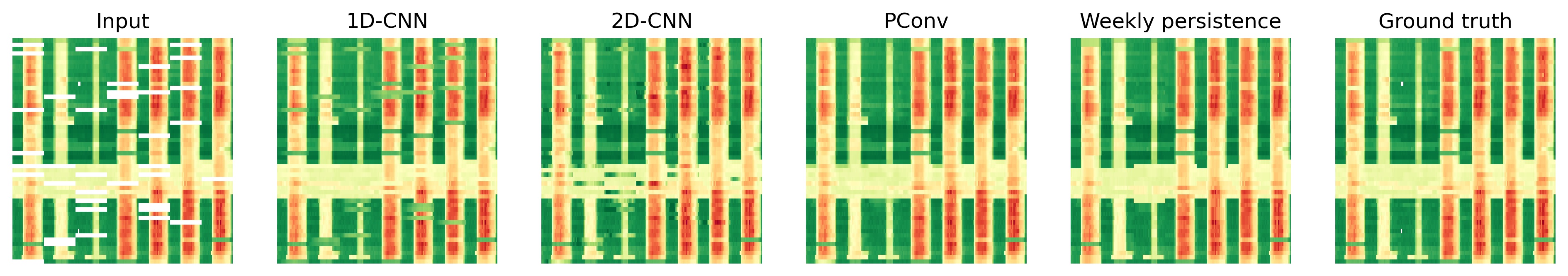}
    %\label{fig:first}
\end{subfigure}

\begin{subfigure}{0.8\textwidth}
    \caption{10\% Random missing (hot water meter)}
    \includegraphics[width=\textwidth]{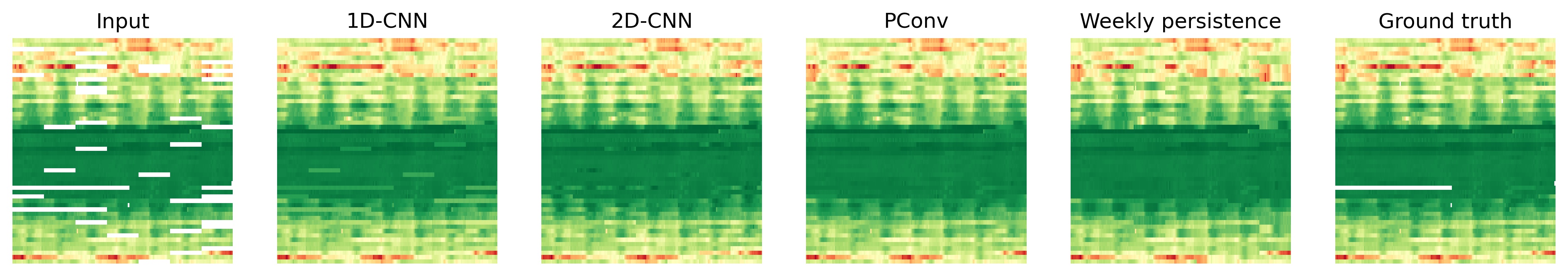}
    %\label{fig:second}
\end{subfigure}

\caption{
    Visual comparisons of models in 2D heatmaps. 
    Each example from left to right: input image, PConv, and ground truth. 
    All images have a $192 \times 192$ pixel dimension.}
\label{fig:heatmap_plots}
\end{figure*}

In the time-series plots of Figure \ref{fig:ts_plots_random} and  \ref{fig:ts_plots_continous}, we look closely at a few examples and compare prediction results across models and missing settings. 
In the case of random missing data shown in plots \ref{fig:ts_plots_c} and \ref{fig:ts_plots_d}, we found that PConv and 2D-CNN, both employing the two-dimensional context, perform remarkably well in the prediction of time-series profiles. 
Conversely, the 1D-CNN fails to capture energy profiles, and the weekly persistence method only offers moderate predictive outcomes. This discrepancy is particularly noticeable in the case of the hot water meter due to its weather dependence.
In \ref{fig:ts_plots_a} and \ref{fig:ts_plots_b}, prediction results with more challenging continuous missing gaps are shown.
Although both PConv and 2D-CNN predictions exhibit stable periodic patterns closely aligned with the original data, the 2D-CNN model struggles to capture sudden shifts in overall trends.
In contrast, PConv and the weekly persistence method are able to account for such trend changes in their imputation predictions.
As for the prediction of hot water usage with continuous missing data, both PConv and 2D-CNN outperform other methods, benefiting from the two-dimensional context. 
The weekly persistence method and 1D-CNN perform poorly in predicting weather-dependent hot water meter, owing to the irregularity in its usage patterns.
Figure \ref{fig:heatmap_plots} illustrates the same results in the form of heatmaps, emphasizing the benefits of a two-dimensional context in the imputation process.

%\begin{figure}[!htb]
%\begin{center}
%\includegraphics[width=1.0\textwidth, trim= 0cm 0cm 0cm 0cm,clip]{Figures/ts_plots.jpg}
%\caption{Time series plots of power meters for comparing imputation results from different methods.}
%\label{fig:ts_plots}
%\end{center}
%\end{figure}

% \newpage
% The heatmaps in Figure \ref{fig:heatmap_plots} illustrate the imputation results of the models on a two-dimensional plane, making it easier to understand the relationship between the prediction results and the context in two dimensions.
% In the prediction results for random missing in (a) and (b), except for 1D-CNN, all other models can predict these randomly scattered daily missing quite well. 
% However, 2D-CNN has a slightly higher error in (a) due to its inability to adapt to time-series shifts. 
% In (c) and (d), both PConv and the weekly persistence model can effectively adjust the prediction results according to the trend shift, making them more accurate than 1D- and 2D-CNN. 
% In the heatmaps of PConv prediction, it can be seen that the prediction trend smoothly change according to the trend shift, while the weekly persistence method appears to change more sharply due to its practice of copying values.

%\begin{figure}[!htb]
%\begin{center}
%\includegraphics[width=1.0\textwidth, trim= 0cm 0cm 0cm 0cm,clip]{Figures/heatmap_plots.jpg}
%\caption{
%    Visual comparisons of models in 2D heatmaps. 
%    Each example from left to right: input image, PConv, and ground truth. 
%    All images have a $192 \times 192$ pixel dimension.}
%\label{fig:heatmap_plots}
%\end{center}
%\end{figure}

\begin{figure*}[!htb]
\centering

\begin{subfigure}{0.8\textwidth}
    \caption{}
    \includegraphics[width=\textwidth]{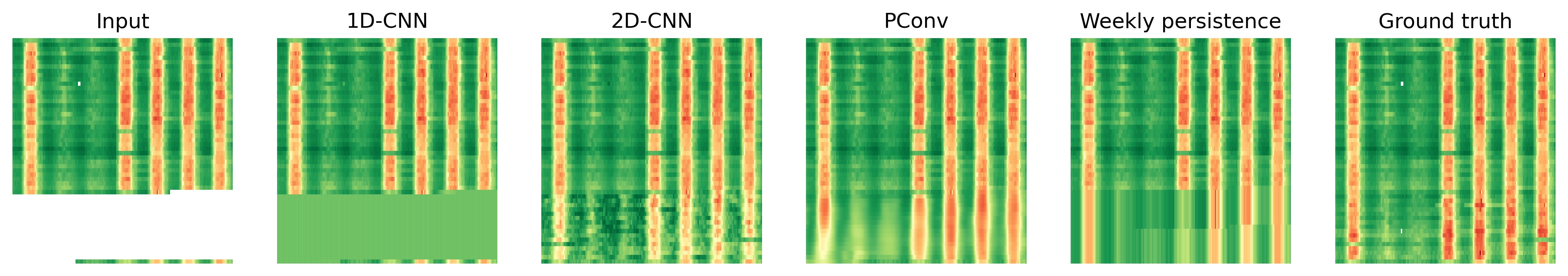}
    \label{fig:fail_plots_a}
\end{subfigure}
\vspace{-0.3cm}

\begin{subfigure}{0.8\textwidth}
    \caption{}
    \includegraphics[width=\textwidth]{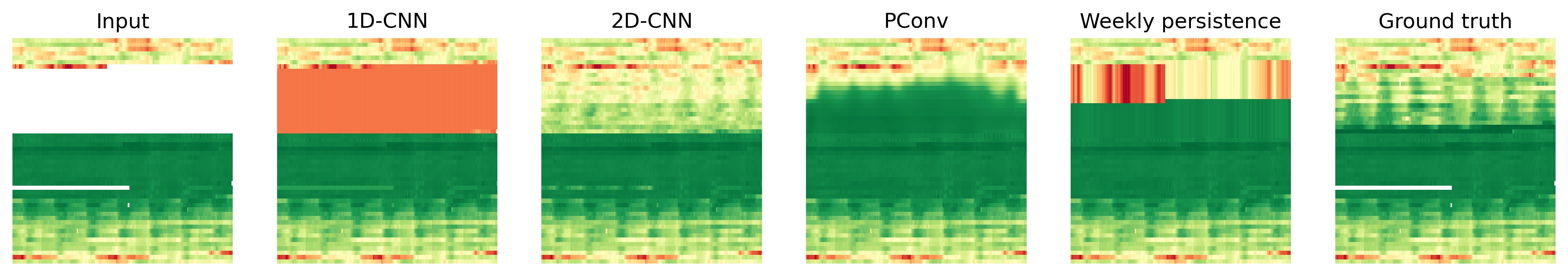}
    \label{fig:fail_plots_b}
\end{subfigure}

\caption{Some examples of failed prediction with high continuous missing rate.}
\label{fig:fail_plots}
\end{figure*}

\newpage
Despite the promising results shown previously, there were instances where the predicted values deviated significantly from the actual values, particularly when the missing rate escalates to 30\%, as illustrated in Figure \ref{fig:fail_plots}.
In Figure \ref{fig:fail_plots_a}, 1D-CNN and the weekly persistence method fail to capture the weekly regularity of energy data, resulting in constant values appearing uniformly colored. Meanwhile, the predictions generated by the 2D-CNN appear to be noisy.
Even though PConv outperforms the other models considerably, it still shows unclear areas, likely resulting from the lack of additional weather information when attempting to impute such long-term missing data.
The difficulties become even more pronounced in Figure \ref{fig:fail_plots_b}, which features 30\% continuous missing data in the hot water meter.
None of the models are able to generate results of satisfactory quality. The absence of weather information, which is particularly crucial for weather-dependent meters, significantly hampers the predictive capabilities of the models.
These failures demonstrate the limitations of imputation models when confronted with long-term continuous missing gaps.

%\begin{figure}[!htb]
%\begin{center}
%\includegraphics[width=1.0\textwidth, trim= 0cm 0cm 0cm 0cm,clip]{Figures/fail_plots.jpg}
%\caption{Some examples of failed prediction with high values of continuous missing rate.}
%\label{fig:fail_plots}
%\end{center}
%\end{figure}

\newpage
Overall, the order of performance from best to worst is PConv, 2D-CNN, weekly persistence, and 1D-CNN. 
PConv, a modern framework for image inpainting with deep learning, consistently performs significantly better than the other models in any task of missing imputation. 
Among the other models, 2D-CNN marginally outperforms the weekly persistence model, whereas 1D-CNN significantly lags behind. 
However, the effectiveness of 2D-CNN and weekly persistence can fluctuate depending on the specific circumstances. 
In situations where there's a trend shift within the missing gaps, weekly persistence might perform slightly better than 2D-CNN as it can incorporate this trend shift into its predictions by replicating the values from the adjacent week.
Furthermore, when the prediction target is to predict missing gaps for weather-dependent meters, such as hot water meters, 2D-CNN performs better than the weekly persistence model. The latter primarily depends on the regularity of the time series, which can be a limiting factor in such cases.

%\begin{table}
%\centering
%\resizebox{0.95\textwidth}{!}{%
%\begin{tabular}{llllllllll}
% & \multicolumn{4}{c}{\textbf{Continuous}} &  & \multicolumn{4}{c}{\textbf{Random}} \\ \cline{2-5} \cline{7-10} 
%\textbf{\begin{tabular}[c]{@{}l@{}}Meter type\end{tabular}} & \textbf{1D-CNN} & \textbf{2D-CNN} & \textbf{PConv} & \textbf{weekly persistence} &  & \textbf{1D-CNN} & \textbf{2D-CNN} & \textbf{PConv} & \textbf{weekly persistence} \\ \hline
%electricity & 0.037162 & 0.677247 & 0.734427 & 0.668361 &  & 0.512474 & 0.783555 & 0.906557 & 0.763082 \\
%chilledwater & 0.038072 & 0.587394 & 0.637105 & 0.507094 &  & 0.733065 & 0.826315 & 0.906382 & 0.757628 \\
%steam & 0.017415 & 0.296295 & 0.411478 & 0.321013 &  & 0.675876 & 0.697772 & 0.826002 & 0.587975 \\
%hotwater & 0.015757 & 0.299769 & 0.40789 & 0.2888 &  & 0.705914 & 0.695937 & 0.846812 & 0.625827 \\ \hline
%\end{tabular}
%}
%\caption{mse meterType table}
%\label{tab:mse_meterType_table}
%\end{table}

\section{Discussion}
\label{sec:discussion}
In this study, we explored the use of various machine learning models, including advanced image inpainting techniques, to impute missing data from energy meters at different missing rates in both random and continuous fashion. 

\subsection{Evaluation and comparison of imputation models}
This study evaluated a range of imputation models, from a na\"ive baseline (i.g., weekly persistence model) to advanced ones (2D-CNN and PConv). 
Notably, this study introduced a unique aspect to the field by incorporating advanced image inpainting techniques, specifically PConv, for imputing missing data. 
Contrary to the majority of previous works that concentrated on imputing short-term missing data (e.g., hourly or daily), our study broadened its scope to cover longer-term periods, such as weekly or even monthly missing data.
The relative ease of imputing short-term missing data based on nearby values was affirmed in the study, thus differentiating our work owing to its long-term focus.
Moreover, the superior performance of PConv over other methods underlined its efficacy, especially when faced with the challenging task of long-term imputation.

Furthermore, most past studies were benchmarked on a restricted number of meters within specific buildings or sites, while the proposed method in this study was comprehensively validated on a benchmark dataset with thousands of meters worldwide.
Consequently, the distinct nature of this study's tasks involving long-term missing data and the inherent complexities of the dataset make it less directly comparable to the imputation results of earlier studies.
Nonetheless, to maintain methodological continuity with past studies, our research incorporated some common deep learning frameworks, such as the CNN-based autoencoder and na\"ive weekly persistence model, as benchmarking references. 
The inclusion of these model frameworks, various missing periods, and our benchmark dataset ensures that our deep learning method based on reshaped data contributes to advancing data imputation in the field.

\subsection{Exploring the maximum time period for imputing continuously missing data}
This study assessed the feasibility and limitation of employing energy time series as images for imputing missing data.
Our results demonstrate that, while the proposed approach PConv outperformed other models via learning context within images, it struggled with the prediction of extended periods of continuous missing values. 
Specifically, when the missing rate of continuous data exceeded 10\%, approximating to about a month of missing, the average R\textsuperscript{2} of PConv dropped below 0.7. 
Additionally, the imputation performance of PConv significantly deteriorated when applied to weather-dependent meters, primarily due to less predictable regularity compared to electricity meters. 
These findings reveal that, even though image-based deep learning models may improve the performance of imputation, their capability to accurately forecast missing values over extended periods remains restricted.

\subsection{Challenges and future direction in imputing data missing}
Imputing continuous missing data, especially with increasing missing rates, poses significant challenges due to the lack of adjacent context values required for long-term prediction. 
This study demonstrates that the advanced deep learning frameworks in computer vision, such as 2D-CNN and PConv, exhibit superior performance via learning context from time-series data represented as images.
However, even with these advanced methods, predicting long-term continuous missing values remains problematic. 
This is evident in the performance of PConv, which noticeably deteriorates when the proportion of continuous missing data exceeds 10\%.

This research emphasizes the necessity for comprehensive context information, such as weather and occupant-related data, to enhance the performance of the imputation model. 
Without these contextual data, the model may struggle to generate accurate predictions for long-term missing values, given that such predictions are typically influenced by weather patterns and human activity.
Encouragingly, future work in this area could benefit from strategies employed in image inpainting studies that integrated additional layers of data to improve accuracy. 
For instance, one study integrated a texture information layer into the imputation process, successfully enhancing the performance by including detailed outlines of human faces or the sketch drawings of buildings \cite{guo2021image}.
Applying the same concept to building energy data imputation, weather and occupant behavior data could serve as valuable additional data layers to improve predictions. 
As an example, the utility of Google Trends data as a human behavior proxy to enhance energy prediction has been proven \cite{fu2022using}. 
Thus, incorporating such additional data layers, human behavior data and weather data, into data imputation for building energy data presents a promising direction for future research.
This approach could potentially address the current challenges in imputing long-term missing data and contribute to more accurate and reliable predictions.

\section{Conclusion}
\label{sec:conclusion}

This study aimed to investigate whether the missing imputation of energy data could be improved by integrating an extra dimension through reshaping energy time series. 
The prediction performance was extensively evaluated by testing various model frameworks and diverse missing rates and types. 
The results revealed the beneficial role of the two-dimensional energy data structure in data imputation, particularly when used with advanced image-based deep learning models like PConv, leading to better imputation performance than classical CNN models and the weekly persistence method.
Even when applied to subsets of different meter types, PConv is able to maintain consistent high prediction accuracy across different meter types. 
However, a challenge arose when PConv attempted to accurately impute long-term continuous missing values beyond a 10\% missing data rate. This situation caused the average R\textsuperscript{2} of PConv to decrease below 0.7, thereby revealing a limitation of the proposed imputation model and reshaped energy data. 

To the best of our knowledge, this is the first study to extensively verify this imputation method using more than one thousand power meters in the field of building energy. 
This research demonstrates the potential of employing two-dimensional reshaped energy data in conjunction with image-based deep learning techniques for imputing missing energy data. 
The proposed method, which has been thoroughly tested and validated on a wide-ranging benchmark dataset with diverse meter types and sources, is scalable and generalizable. 
With the future availability of more energy datasets, our approach holds promise as a potent tool for automatically imputing missing data on a significant scale. 
This could offer substantial benefits to the industry by facilitating enhanced energy predictions and management based on the imputed data. Furthermore, the proposed method could find wider application across other time series data in the built environment, such as HVAC, lighting, and appliances data, offering a holistic benefit to the building industry.

\section{Limitations}
While this study yielded exciting results and made significant contributions, there are several limitations that should be acknowledged and to be addressed in the future. 
The first limitation is that our focus was on imputing continuous and random missing gaps in separate tasks, neglecting other forms of missing data patterns encountered in the real world. 
For example, multiple scattered continuous missing gaps or a combination of both random and continuous missing were not specifically considered and assessed in this study.
Secondly, the study did not consider the integration of weather data, despite its well-established impact on energy consumption.
Their inclusion in the imputation models could provide valuable contextual information for more accurate predictions.
Similarly, the integration of calendar data, which captures human events like national holidays or extended vacations that influence energy consumption, could further enhance the imputation process. 
Investigating the integration of these contextual factors would contribute to more comprehensive and accurate imputation models.
These limitations suggest that further research is needed to fully explore and leverage the potential of image-based techniques for imputing missing energy data.

\section{Reproducibility}
This analysis can be reproduced using the data and code from the following GitHub repository: \url{https://github.com/buds-lab/Filling-time-series-gaps-using-image-techniques}. 

\section*{CRediT author statement}
\textbf{CF}: Conceptualization, Methodology, Software, Formal Analysis, Investigation, Data Curation, Visualization, Writing - Original Draft; \textbf{MQ}: Methodology, Writing - Reviewing \& Editing; \textbf{ZN}: Methodology, Writing - Reviewing \& Editing; \textbf{CM}: Conceptualization, Methodology, Resources, Writing - Reviewing \& Editing, Supervision, Project administration, Funding acquisition.

\section*{Funding}
This research is funded by the NUS-based Singapore MOE Tier 1 Grant titled Ecological Momentary Assessment (EMA) for Built Environment Research (A-0008301-01-00).

\section*{Acknowledgements}

\bibliographystyle{model1-num-names}
\bibliography{mybibfile}

\end{document}